%% file: arxiv.tex
\documentclass{article}

\usepackage[preprint]{neurips_2025}

\usepackage[utf8]{inputenc} 
\usepackage[T1]{fontenc}    
\usepackage{hyperref}       
\usepackage{url}            
\usepackage{booktabs}       
\usepackage{amsfonts}       
\usepackage{nicefrac}       
\usepackage{microtype}      
\usepackage{xcolor}         
\usepackage{graphicx}
\usepackage[most]{tcolorbox}
\usepackage{tabularx}
\usepackage{placeins}
\usepackage{float}
\usepackage{subcaption}
\usepackage{geometry}

\newcommand\xinyu[1]{\textcolor{blue}{Xinyu: #1}}
\input{macros}

\title{\datasetnamelogo: Can Foundation Models Understand Multimodal Gridded Geo-Spatial Data?}

\author{%
    Bowen Jiang\textsuperscript{1, 2} \quad
    Yangxinyu Xie\textsuperscript{1, 2} \quad
    Xiaomeng Wang\textsuperscript{1} \quad
    Jiashu He\textsuperscript{1} \quad
    Joshua Bergerson\textsuperscript{2} \\
    \textbf{
    John K. Hutchison\textsuperscript{2}  \quad
    Jordan Branham \textsuperscript{2}  \quad
    Camillo J. Taylor\textsuperscript{1} \quad
    Tanwi Mallick\textsuperscript{2}
    }\\
    \begin{tabular}{cc}
        \begin{tabular}{c}
            University of Pennsylvania\textsuperscript{1} \\
            Philadelphia, PA, 19104
        \end{tabular}
        &
        \begin{tabular}{c}
            Argonne National Laboratory\textsuperscript{2} \\
            Lemont, IL, 60439
        \end{tabular}
    \end{tabular} \\
    \small{\texttt{\{bwjiang@seas, xinyux@wharton, cjtaylor@seas\}.upenn.edu, tmallick@anl.gov}}
}

\begin{document}

\maketitle

\input{sections/abstract.tex}

\input{sections/intro}
\input{sections/overview}

\input{sections/experiment_arxiv}
\input{sections/related_work}
\input{sections/conclusion}

\section*{Acknowledgment}

This work was supported by Laboratory Directed Research and Development funding from Argonne National Laboratory, provided by the Office of Science, U.S. Department of Energy under Contract No. DE-AC02-06CH11357.
This work also utilized resources provided by the Argonne Leadership Computing Facility at Argonne National Laboratory.
C. J. Taylor acknowledge support from National Science Foundation grant CCF-2112665 (TILOS).
The funders had no role in study design, data collection, data analysis or interpretation, or manuscript preparation.

\bibliographystyle{ref_style}
\bibliography{refs}

\input{sections/appendix}

\end{document}

%% file: macros.tex
\newcommand{\datasetnamelogo}{\includegraphics[height=1em]{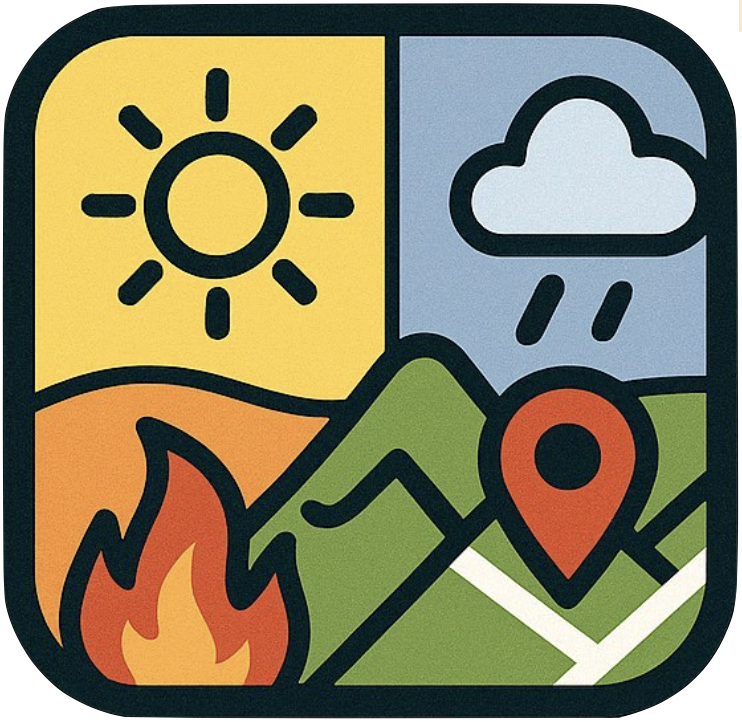}~GeoGrid-Bench}
\newcommand{\datasetname}{GeoGrid-Bench}

\definecolor{color_boxback}{RGB}{245,245,245}
\definecolor{color_boxframe}{RGB}{45,45,45}
\definecolor{color_overall}{RGB}{236, 213, 216}
\definecolor{color_location}{RGB}{188, 227, 148}
\definecolor{color_time}{RGB}{243, 204, 140}
\definecolor{color_climate}{RGB}{176, 217, 219}

\tcbset{
  aibox/.style={
    width=1\linewidth,
    top=7pt,
    bottom=5pt,
    colback=blue!6!white,
    colframe=black,
    colbacktitle=black,
    enhanced,
    center,
    attach boxed title to top left={yshift=-0.1in,xshift=0.15in},
    boxed title style={boxrule=0pt,colframe=white,},
    float,
    floatplacement=t,
  }
}
\newtcolorbox{AIbox}[2][]{aibox,title=#2,#1}

\definecolor{codegray}{gray}{0.95}

%% file: sections/abstract.tex
\begin{abstract}

  We present \datasetname, a benchmark designed to evaluate the ability of foundation models to understand geo-spatial data in the grid structure. Geo-spatial datasets pose distinct challenges due to their dense numerical values, strong spatial and temporal dependencies, and unique multimodal representations including tabular data, heatmaps, and geographic visualizations. To assess how foundation models can support scientific research in this domain, \datasetname~features large-scale, real-world data covering 16 climate variables across 150 locations and extended time frames. The benchmark includes approximately 3,200 question-answer pairs, systematically generated from 8 domain expert-curated templates to reflect practical tasks encountered by human scientists. These range from basic queries at a single location and time to complex spatiotemporal comparisons across regions and periods. Our evaluation reveals that vision-language models perform best overall, and we provide a fine-grained analysis of the strengths and limitations of different foundation models in different geo-spatial tasks. This benchmark offers clearer insights into how foundation models can be effectively applied to geo-spatial data analysis and used to support scientific research.\footnote{All code and data are publicly available at our Github repository \url{https://github.com/bowen-upenn/GeoGrid_Bench} and Huggingface \url{https://huggingface.co/datasets/bowen-upenn/GeoGrid_Bench}.}
\end{abstract}

%% file: sections/intro.tex
\section{Introduction}

\begin{figure}[t]
  \centering
  \includegraphics[width=0.9\linewidth]{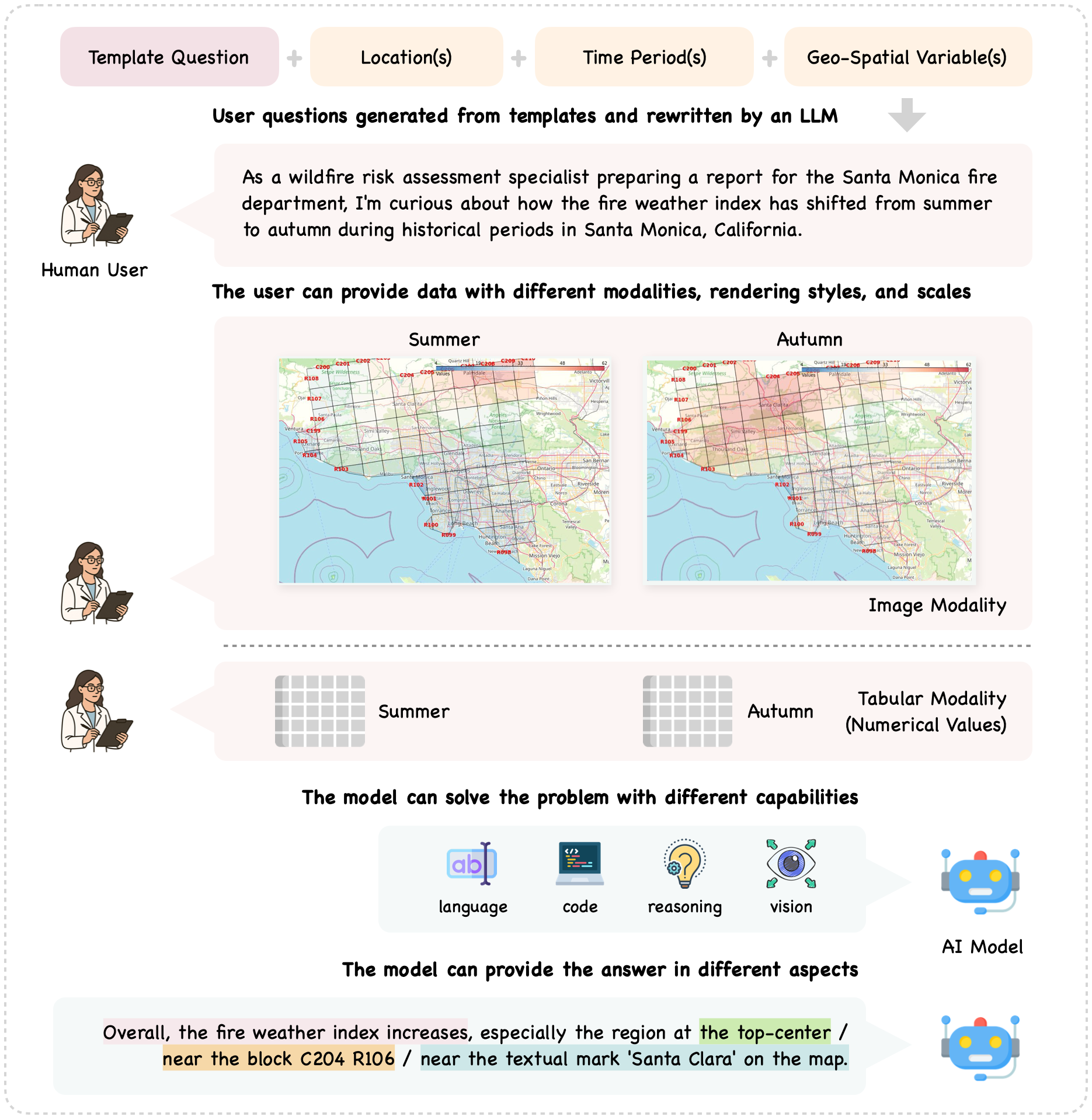}
  \caption{\textbf{Overview of \datasetname.} The benchmark features questions generated from templates that vary by location, time period, and climate variable, then rewritten with natural language context. Each question is paired with multimodal input—either heatmaps as images or tabular grids of numerical values. We evaluate models on their ability to solve the queries through different modalities—natural language, code, or vision. Ground-truth answers capture find-grained aspects like \textcolor{color_overall}{overall trends}, \textcolor{color_location}{spatial references} (from top-left to lower-right), \textcolor{color_time}{coordinate references} (row and column indices), and \textcolor{color_climate}{label references} (textual marks on the maps), whenever available.}
  \vspace{-2mm}
  \label{fig:overview}
\end{figure}

Foundation models have demonstrated transformative capabilities across diverse domains, ranging from language and vision to programming and reasoning~\citep{hurst2024gpt, jaech2024openai, jiang2024towards, jiang2024multi, jiang2024peek, balachandran2024eureka, jiang2024survey, he2024give}. Their rapid advancement has naturally inspired research exploring their utility in scientific contexts, particularly in critical fields like climate science and natural hazard assessment~\citep{mai2022towards, xie2024wildfiregpt, xie2025ragbasedmultiagentllmnatural, nguyen2023climax, mai2023opportunities, de2025information, mallick2025understanding}, where accurate, data-intensive decision-making can profoundly impact human well-being.

Geo-spatial data pose distinct challenges for foundation models due to their inherent spatio-temporal dependencies and exceptionally high data density. Unlike typical tabular records for knowledge retrieval~\citep{zhang2023tablellama, pasupat2015compositional, zhang2025survey} or natural images, climate data exists in structured, gridded formats with complex, interconnected numerical values often represented through modalities such as tables, heatmaps, or geographic images spanning across space and time. These data are typically organized in highly structured, gridded formats that encode interconnected numerical values across spatial and temporal dimensions. Each data point is not an isolated unit but part of a dense, multi-dimensional array that reflects physical processes, environmental interactions, or geographical phenomena evolving over time. Meanwhile, models can also easily get lost in the context~\citep{liu2023lost} with overwhelming volumes of values per sample.

Informed decision-making in fields such as disaster response, climate science, and urban development depends on the ability to detect and interpret patterns across regions and over time. However, there remains a lack of benchmarks that directly address the unique challenges posed by geo-spatial gridded data. Most existing efforts docus on object detection, semantic segmentation, object counting, captioning, or scene understanding of Earth observation images~\citep{lacoste2023geo, danish2024geobench, zhang2024good, zheng2023segment, wang2024earthvqa, muhtar2024lhrs, bazi2024rs, kuckreja2024geochat}, function calls to the Geographic Information System (GIS) or SQL queries for data retrieval~\citep{krechetova2025geobenchx, jiang2024chatgpt, ning2025autonomous, mooney2023towards, zhang2023geogpt}, or simplified query setups that overlook the spatial-temporal complexities in practical geo-spatial analysis~\citep{bhandari2023large}.

To understand how foundation models can assist geo-spatial data analysis, we introduce \datasetname, a benchmark explicitly designed to evaluate model performance on multimodal, real-world geo-spatial data. We adopt domain expert-curated query templates to reflect realistic questions that practitioners would encounter in geo-spatial analysis—providing data in both tabular and image formats. These tasks range from simple queries about a fixed location and time to more complex analyses involving multiple locations and temporal comparisons. For each template, we develop oracle code that is applied uniformly to all query instances, enabling scalable and consistent generation of question-answer pairs.
Our contributions can be summarized as follows:
\begin{itemize}
    \item \textbf{Large-scale, real-world data:} A domain-centric benchmark built on large-scale, real-world climate projection data, presented in multimodal formats commonly used by actual practitioners, including structured numerical tables and geographic visualizations.

    \item \textbf{Scalable query generation:} A systematic user query generation pipeline based on domain expert-designed templates, reflecting diverse and realistic scientific challenges.

    \item \textbf{Comprehensive evaluation:} Evaluation of foundation models with language, coding, multimodal, and reasoning capabilities across find-grained answer aspects and data modalities to diagnose their strengths and weaknesses in geo-spatial analysis tasks.

\end{itemize}

Through comprehensive evaluations, we find that visualizing dense, gridded geo-spatial data as heatmaps is the most accessible format for existing foundation models to interpret. In contrast, models struggle to generate flawless code for completing these tasks. Across all model types, identifying broad trends proves easier than making fine-grained regional distinctions, and models exhibit varying strengths and weaknesses depending on the task. With \datasetname, we aim to shed light on the strengths and limitations of current foundation models when applied to multimodal geo-spatial data, a core yet underexplored format in climate science. Our goal is to support and advance the development of practical AI-assisted tools that can aid scientific research and decision-making.


\begin{figure}[t]
  \centering
  \includegraphics[width=1\linewidth]{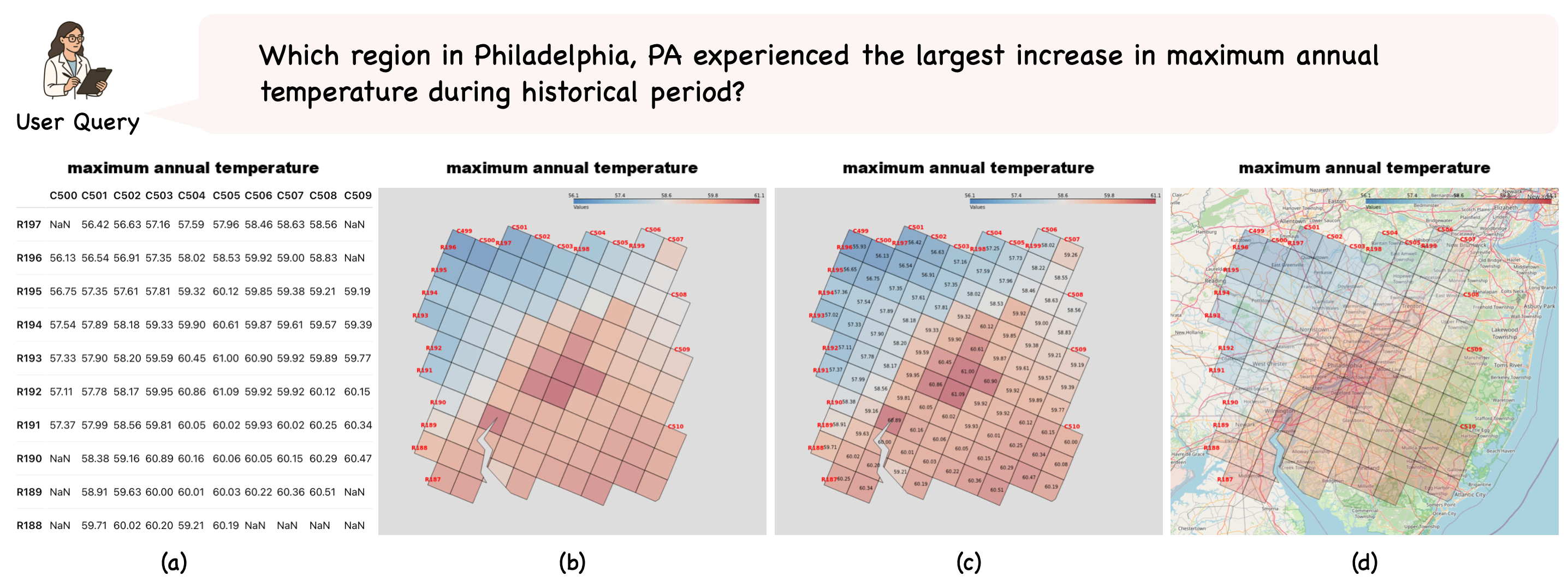}
  \vspace{-5mm}
  \caption{We prepare every data sample in one of the four formats: (a) 2D table as a textual string. (b) standalone heatmap; (c) heatmap with overlaid numerical annotations at each grid cell; (d) heatmap overlaid on an actual geographic base map. These formats reflect real-world climate data practices and differ markedly from typical natural images seen by foundation models. More in Appendix~\ref{sec:more_vis}.}
  \label{fig:image_formats}
  \vspace{-2mm}
\end{figure}

%% file: sections/overview.tex
\section{\datasetnamelogo: Overview of Data Features and Tasks}

\label{sec: Overview}

\datasetname~aims to reflect the real-world challenges that scientists face when analyzing geo-spatial data at scale. To achieve this, it features \textit{large-scale, real-world} geo-spatial data sourced and sampled from ClimRR~\citep{climrr2023}, capturing the complexity of environmental conditions across 150 locations in North America. The grid spans 16 diverse climate variables, such as temperature extremes, precipitation, wind speeds, humidity, fire weather indices, and degree days. An overview of user-model interaction is shown in Figure~\ref{fig:overview}. 

\textit{\datasetname~is built to capture the unique grid structure.} Climate projection data are typically organized across spatial grids and time sequences, resulting in dense, high-dimensional arrays. The data is inherently interconnected, with each point influenced by its geographic neighbors and historical context. This structure poses unique challenges: models must capture spatio-temporal dependencies and handle variability across scales to derive meaningful insights.

\textit{Geo-spatial data is also inherently multimodal, presented as tabular data, heatmaps, or geographic visualizations}, with each format sharing alignment across a spatial grid structure. Each grid cell encodes a rich array of numerical data that captures localized atmospheric behavior and climate dynamics over time. This multimodal grid structure makes our \datasetname~an ideal testbed for foundation models designed to reason across space, time, and modality. To perform well, foundation models must integrate spatial context from neighboring cells, understand temporal trends across multi-year projections, and interpret information presented in diverse formats and patterns. \datasetname~reflects this complexity and we show examples of the data formats in Figure~\ref{fig:image_formats}.

To capture the wide range of questions concerning practitioners at the forefront of geo-spatial analysis, we surveyed 13 domain experts in natural hazard risk domains, resulting in 8 template questions based on their input and around 3200 query instances in \datasetname. Each template includes placeholders based at one or two geographic locations, time frames, and climate variables, requiring one to eight data frames. This design allows us to generate a scalable set of scientifically concrete queries that reflect analytical goals. Specifically, \datasetname~evaluates the following capabilities of foundation models:
\begin{itemize}
    \item \textbf{Identifying regions with the most significant patterns.} This is essential for prioritizing attention and resources in applications like disaster response and climate monitoring, enabling us to detect anomalies or hotspots that require timely intervention or analysis.
    \item \textbf{Comparing data across different locations and times.} This is essential for uncovering spatial disparities, understanding regional dynamics, and tracking changes over time. 
    \item \textbf{Analyzing temporal trends and seasonal variations.} This is essential for practitioners to anticipate recurring patterns and detect long-term changes to make informed decisions. 
    \item \textbf{Interpreting data in multimodal formats.} This is essential for understanding the ability of foundation models to interpret real-world geo-spatial data that is multimodal in nature.
\end{itemize}



\begin{table}[H]
\centering
\small
\renewcommand{\arraystretch}{1.4}  
\begin{tabular}{p{\linewidth}}
\hline
\textbf{Templates that require one data frame} \\
\hline
1. Which region in the \raisebox{0pt}[0.8\height][0.8\depth]{\setlength{\fboxsep}{0pt}\colorbox{color_location}{\strut \{location1\}}} experienced the largest increase in \raisebox{0pt}[0.8\height][0.8\depth]{\setlength{\fboxsep}{0pt}\colorbox{color_climate}{\strut \{variable1\}}} during \raisebox{0pt}[0.8\height][0.8\depth]{\setlength{\fboxsep}{0pt}\colorbox{color_time}{\strut \{time\_frame1\}}}? \\
\hline
\textbf{Templates that require two data frames} \\
\hline
2. How has \raisebox{0pt}[0.8\height][0.8\depth]{\setlength{\fboxsep}{0pt}\colorbox{color_climate}{\strut \{variable1\}}} changed between \raisebox{0pt}[0.8\height][0.8\depth]{\setlength{\fboxsep}{0pt}\colorbox{color_time}{\strut \{time\_frame1\}}} and \raisebox{0pt}[0.8\height][0.8\depth]{\setlength{\fboxsep}{0pt}\colorbox{color_time}{\strut \{time\_frame2\}}} in the \raisebox{0pt}[0.8\height][0.8\depth]{\setlength{\fboxsep}{0pt}\colorbox{color_location}{\strut \{location1\}}}? \\
\hline
3. What is the correlation between \raisebox{0pt}[0.8\height][0.8\depth]{\setlength{\fboxsep}{0pt}\colorbox{color_climate}{\strut \{variable1\}}} and \raisebox{0pt}[0.8\height][0.8\depth]{\setlength{\fboxsep}{0pt}\colorbox{color_climate}{\strut \{variable2\}}} in the \raisebox{0pt}[0.8\height][0.8\depth]{\setlength{\fboxsep}{0pt}\colorbox{color_location}{\strut \{location1\}}} during \raisebox{0pt}[0.8\height][0.8\depth]{\setlength{\fboxsep}{0pt}\colorbox{color_time}{\strut \{time\_frame1\}}}? \\
\hline
4. How does \raisebox{0pt}[0.8\height][0.8\depth]{\setlength{\fboxsep}{0pt}\colorbox{color_climate}{\strut \{variable1\}}} compare between \raisebox{0pt}[0.8\height][0.8\depth]{\setlength{\fboxsep}{0pt}\colorbox{color_location}{\strut \{location1\}}} and \raisebox{0pt}[0.8\height][0.8\depth]{\setlength{\fboxsep}{0pt}\colorbox{color_location}{\strut \{location2\}}} during \raisebox{0pt}[0.8\height][0.8\depth]{\setlength{\fboxsep}{0pt}\colorbox{color_time}{\strut \{time\_frame1\}}}? \\
\hline
\textbf{Templates that require four data frames }\\
\hline
5. What is the \textit{seasonal} variation of \raisebox{0pt}[0.8\height][0.8\depth]{\setlength{\fboxsep}{0pt}\colorbox{color_climate}{\strut \{climate\_variable1\}}} in \raisebox{0pt}[0.8\height][0.8\depth]{\setlength{\fboxsep}{0pt}\colorbox{color_location}{\strut \{location1\}}} during \raisebox{0pt}[0.8\height][0.8\depth]{\setlength{\fboxsep}{0pt}\colorbox{color_time}{\strut \{time\_frame1\}}}? \\
\hline
6. Which \textit{season} in \raisebox{0pt}[0.8\height][0.8\depth]{\setlength{\fboxsep}{0pt}\colorbox{color_time}{\strut \{time\_frame1\}}} saw the highest levels of \raisebox{0pt}[0.8\height][0.8\depth]{\setlength{\fboxsep}{0pt}\colorbox{color_climate}{\strut \{variable1\}}} in \raisebox{0pt}[0.8\height][0.8\depth]{\setlength{\fboxsep}{0pt}\colorbox{color_location}{\strut \{location1\}}}? \\
\hline
7. Which of \raisebox{0pt}[0.8\height][0.8\depth]{\setlength{\fboxsep}{0pt}\colorbox{color_location}{\strut \{location1\}}} or \raisebox{0pt}[0.8\height][0.8\depth]{\setlength{\fboxsep}{0pt}\colorbox{color_location}{\strut \{location2\}}} experienced a greater change in \raisebox{0pt}[0.8\height][0.8\depth]{\setlength{\fboxsep}{0pt}\colorbox{color_climate}{\strut \{variable1\}}} throughout \raisebox{0pt}[0.8\height][0.8\depth]{\setlength{\fboxsep}{0pt}\colorbox{color_time}{\strut \{time\_frame1\}}} and \raisebox{0pt}[0.8\height][0.8\depth]{\setlength{\fboxsep}{0pt}\colorbox{color_time}{\strut \{time\_frame2\}}}? \\
\hline
\textbf{Templates that require eight data frames }\\
\hline
8. How does the \textit{seasonal} variation of \raisebox{0pt}[0.8\height][0.8\depth]{\setlength{\fboxsep}{0pt}\colorbox{color_climate}{\strut \{variable1\}}} in \raisebox{0pt}[0.8\height][0.8\depth]{\setlength{\fboxsep}{0pt}\colorbox{color_location}{\strut \{location1\}}} compare to that in \raisebox{0pt}[0.8\height][0.8\depth]{\setlength{\fboxsep}{0pt}\colorbox{color_location}{\strut \{location2\}}} for \raisebox{0pt}[0.8\height][0.8\depth]{\setlength{\fboxsep}{0pt}\colorbox{color_time}{\strut \{time\_frame1\}}}? \\
\hline
\end{tabular}
\vspace{3mm}
\caption{\textbf{Template questions in \datasetname.} We develop those questions with domain experts. Each question includes placeholders for one or two locations, time frames, and geo-spatial variables. This design enables scalable question construction while capturing varying levels of complexity based on the number of data frames involved.}
\label{tab:template_questions}
\vspace{-3mm}
\end{table}

\begin{figure}[H]
\centering
\small
\begin{tcolorbox}[title=Full List of Climate Variables in \datasetname,
colback=blue!6!white,
colframe=color_boxframe]
Maximum Annual Temperature, Minimum Annual Temperature, Consecutive Days with No Precipitation, Cooling Degree Days, Fire Weather Index, Maximum Daily Heat Index, Maximum Seasonal Heat Index, Number of Days with Daily Heat Index > 95°F/105°F/115°F/125°F, Heating Degree, Annual Total Precipitation, Maximum Seasonal Temperature, Minimum Seasonal Temperature, Wind Speed.
\label{fig:climate}
\end{tcolorbox}
\end{figure}

\begin{figure}[H]
\centering
\small
\begin{tcolorbox}[title=Summary Of Data Statistics in \datasetname,
colback=blue!6!white,
colframe=color_boxframe]
\vspace{-1mm}
\datasetnamelogo~comprises 3,200 user queries, 16 climate variables, and 150 unique locations in North America, with 50 locations associated with each climate variable. It includes 8 query templates and each template has 100 filled query instances, which can be easily scaled up. Each query instance has 4 versions, including data in either tabular or one of three image formats. Each query instance requires between 1 and 8 data frames, each covering up to around 150 grid cells or an edge size of 144 km on maps. The data spans periods from 1995 to projected values toward the end of the 21st century.
\end{tcolorbox}
\label{fig:summary}
\end{figure}
\vspace{-6mm}

\section{Constructing \datasetnamelogo~At Scale}
\label{sec: constructing}
\vspace{-2mm}
\begin{figure}[H]
  \centering
  \includegraphics[width=1\linewidth]{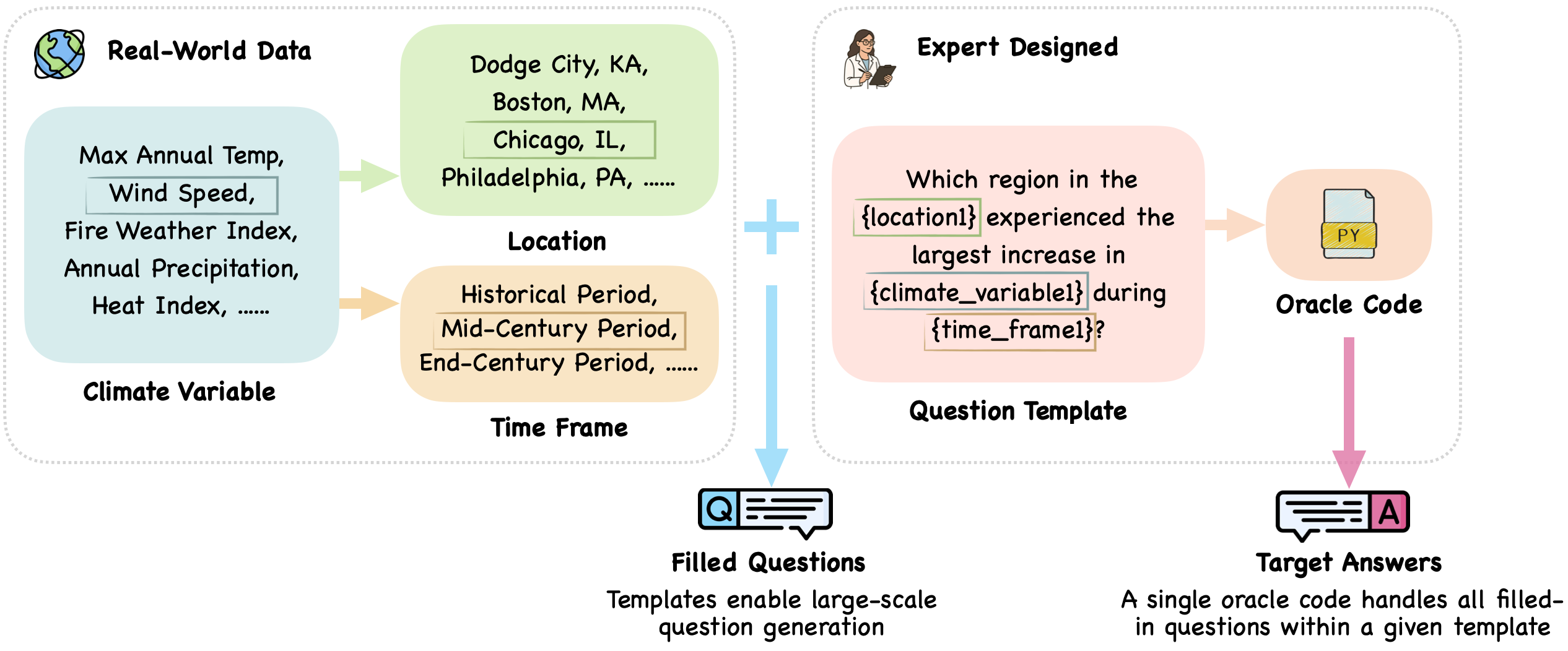}
  \vspace{-5mm}
  \caption{\textbf{Overview of the example curation process.} Each example in \datasetname~is constructed by combining a query template with sampled climate variables, locations, and time frames from real-world climate data. Each template is paired with a corresponding oracle code that deterministically generates target answers for all filled-in question instances under that template.}
  \label{fig:data_curation}
\end{figure}



\subsection{\datasetname~features diverse real-world geo-spatial data}\label{sec:data}
We illustrate our sample curation process in Figure~\ref{fig:data_curation}. Each data sample is formed by extracting a specific climate-location-time slice from the ClimRR~\citep{climrr2023} dataset. We sample from the 16 climate variables listed above. For each climate variable, we select around 50 locations where this climate variable is the most prominent, resulting in a total of 150 distinct locations across all climate variables, a subset of ClimRR. For example, the benchmark includes more regions in Southern California for wildfire risk, while precipitation-related examples are more concentrated in the Pacific Northwest to reflect region-specific climate concerns.


We render each data sample in either a \textbf{tabular} or \textbf{image} format, both structured over a spatial grid. For a given location and its longitude and latitude, we retrieve all grid cells within a square region with edge size $84$ to $144$ km around it, resulting in approximately 50 to 150 entries in the $12$-by-$12$ km grid. In the \textbf{tabular} modality, we prepare each table with numerical values, a caption, and row and column indices as textual strings. In the \textbf{image} modality, we prepare three types of visualization with increasing information densities, as shown in Figure~\ref{fig:image_formats}:
(1) A standalone heatmap,
(2) A heatmap with overlaid numerical annotations at each grid cell.
and (3) A heatmap overlaid on an actual geographic base map.
Specifically, we render the tabular data as a heatmap with color gradients. This heatmap is optionally added with numerical annotation of the value on each cell, or overlaid on a geographic base map~\citep{openstreetmap} using Folium~\citep{folium}. To maintain consistency with the tabular format, we also render row and column indices around the heatmap. This visualization offers a richer representation to mirror common practices in real-world analysis.
To isolate the challenge of data retrieval, \datasetname~provides the foundation model during evaluation with all necessary data frames in either tabular or image formats, focusing solely on whether the model can solve the problem given the relevant information.

\subsection{\datasetname~builds on expert-curated templates for scalable query generation}\label{sec:aspect}

To ensure that the benchmark reflects the types of analysis most relevant to practitioners in geospatial research, we consulted 13 domain experts across multiple scientific disciplines. These experts routinely engage with geo-spatial data to identify patterns, assess risks, and support decision-making under uncertainty. Based on in-depth discussions about the analytical tasks they perform, we develop eight representative question templates. Each template takes as input one or two climate variables, locations, and time frames and outputs a filled-in user query in our benchmark, and may require between one and eight data frames to answer.
This structured approach enables the automatic generation of a wide variety of concrete, data-driven queries tailored to real-world analytical needs.



For every template, we manually craft oracle code that deterministically solves the question and prepares ground-truth answers in desired formats. \textit{Crucially, the same oracle applies uniformly to every query generated from a given template, enabling the scalability of the benchmark. As a result, once a template and its oracle are validated, we ensure the quality of every generated instance.}


Each question is a multiple-choice with four options, all generated by the oracle code rather than a language model. Recognizing that a foundation model may excel at different aspects in answering a geo-spatial query, the benchmark has each query probe a different aspect in giving the answer, as shown in Figure~\ref{fig:overview}. Specifically, answer options target the following aspects:
(1) Overall patterns (e.g., the wildfire risk overall increases).
(2) Spatial references (e.g., the highest wildfire risk occurs around the top-left region).
(3) Coordinate references (e.g., the highest wildfire risk occurs around Column 204 Row 106).
(4) Label references (e.g., the highest wildfire risk occurs near the textual label "Santa Clara" on the map), which is only available for the image type "heatmap overlaid on an actual geographic base map".

In addition, to explore which data modalities most effectively support geo-spatial analysis, we evaluate models across three input settings: \textbf{language-only}, \textbf{language and code}, and \textbf{language and vision}. Detailed prompting strategies for each setting are provided in Appendix~\ref{sec:prompt}.
In each mode, we provide the model with the user query, the relevant data (in either tabular or image format), all four multiple-choice options, and system instructions as inputs. We extract the model's final answer following the special tokens \textit{"\#\#\#\#Final Answer"} to facilitate answer parsing. 
If the model fails to provide an explicit option (a), (b), (c), or (d), we use a sentence embedding model~\citep{reimers-2019-sentence-bert} to identify the most similar option based on the model's response. When the model outputs Python code, we execute the code in a shell environment to extract the final answers.


%% file: sections/experiment_arxiv.tex
\section{Experiment}

\begin{figure}[t]
  \centering
  \includegraphics[width=1\linewidth]{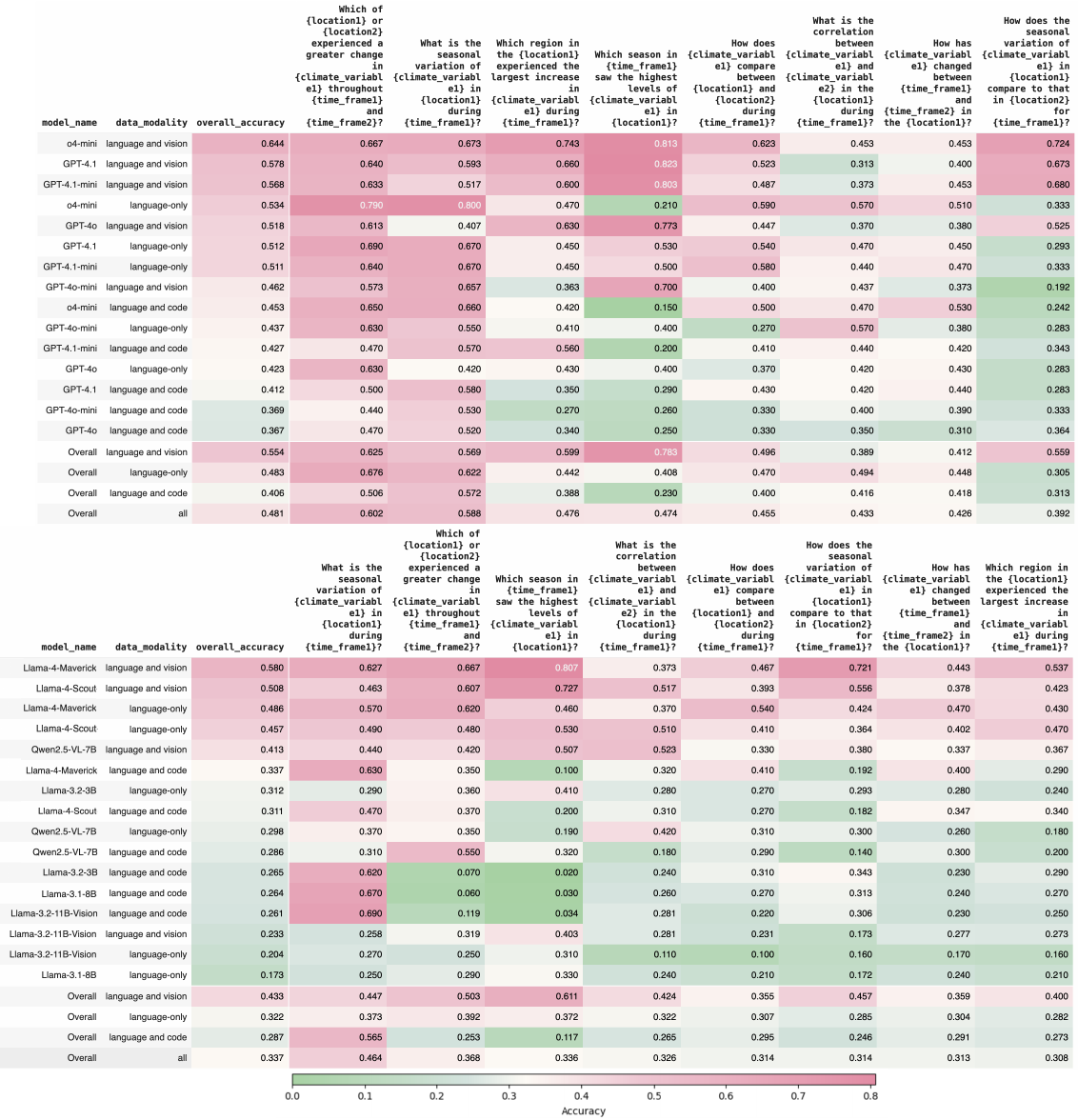}
  \caption{Evaluation results. The top table shows OpenAI models and the bottom table shows open-source models. Each row corresponds to one model with one data modality—language-only, language and code, or language and vision, while each column represents a query template in Table~\ref{tab:template_questions}.}
  \label{fig:results}
\end{figure}

\begin{figure}[t]
  \centering
  \includegraphics[width=0.9\linewidth]{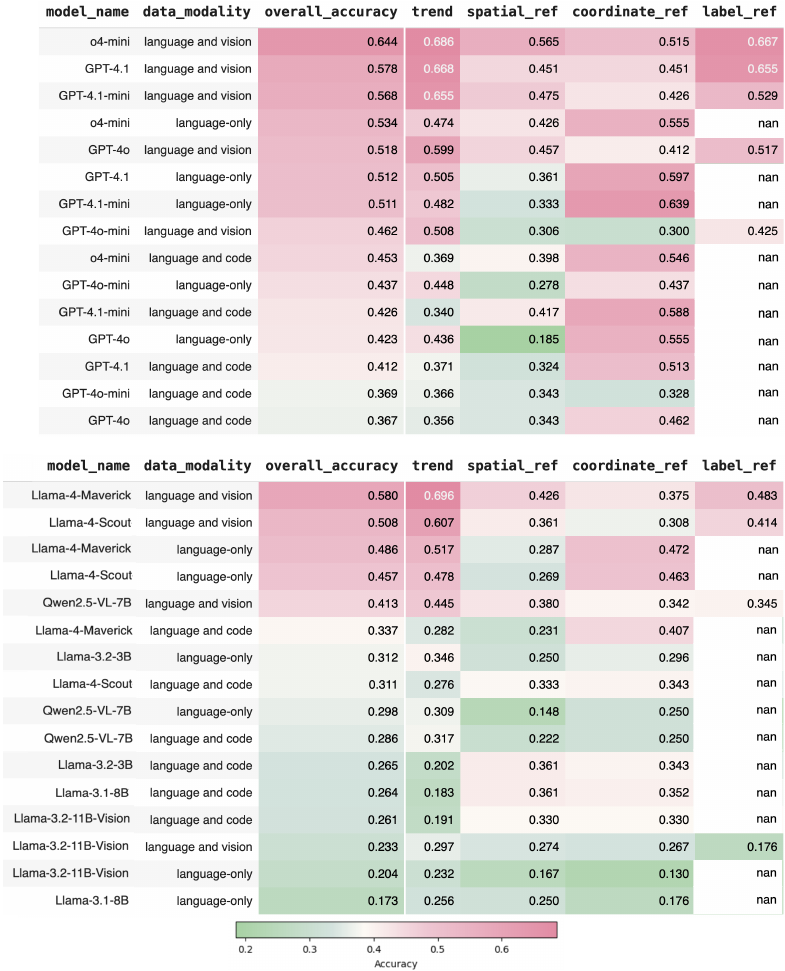}
  \caption{More evaluation results. The top table shows OpenAI models and the bottom table shows open-source models evaluated under different data modalities. Columns represent fine-grained answer aspects defined in Section~\ref{sec:aspect}, including trend, spatial references, coordinate references, and label references. There exist NaN values since the label reference is only available for the vision modality.}
  \label{fig:results2}
  \vspace{-3mm}
\end{figure}

\begin{figure}[t]
  \centering
  \includegraphics[width=0.9\linewidth]{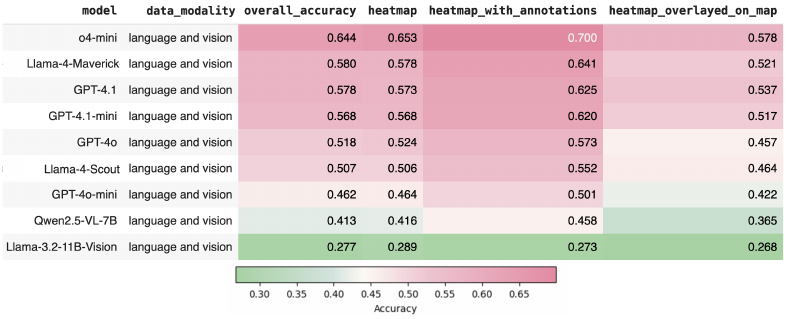}
  \caption{More evaluation results on vision-language models, which are evaluated on three visualization types, as mentioned in Section~\ref{sec:data} and Figure~\ref{fig:image_formats}.}
  \label{fig:results3}
\end{figure}

\subsection{Experimental Setup}

\label{sec: Experimental Setup}

We benchmark a range of state-of-the-art closed-source and open-source models on \datasetname. Our evaluation covers $5$ models from OpenAI, including o4-mini, GPT-4.1, GPT-4.1-mini, GPT-4o, and GPT-4o-mini~\citep{openai2024o4mini, openai2025gpt41, hurst2024gpt}, and $6$ open-source models including Llama-4-Maverick, Llama-4-Scout, Llama-3.2-11B-Vision, Llama-3.2-3B, Llama-3.1-8B~\citep{grattafiori2024llama, meta2024llama4}, and Qwen-2.5-VL-7B~\citep{bai2025qwen2}. OpenAI models are accessed via API calls, and Llama-4 models are accessed through the Lambda Inference API. Inferences for other open-source models run locally on four NVIDIA A100-SXM4 GPUs with 40GB of VRAM. For all models, we set \texttt{max\_new\_tokens} as $1024$ with default temperature and sampling strategies.

\subsection{Evaluation Results and Findings}

\textbf{Vision-language models achieve the strongest performance in geo-spatial tasks}
Among the models we evaluate, o4-mini achieves overall the highest performance, while Llama-4-Maverick leads among open-source models, as shown in Figure~\ref{fig:results}. Overall, models that receive input in the vision modality consistently outperform those using language-only input. This suggests that converting geo-spatial gridded data into heatmap visualizations—rather than presenting models directly with large volumes of raw numerical values in tabular forms—enables foundation models to more effectively interpret such data with complex spatial-temporal patterns.

\textbf{Inferior performance in code highlights the need for more agentic models in geo-spatial tasks}
Contrary to our expectations, foundation models leveraging programming code do not outperform their language-only counterparts on our task. Upon closer inspection, much of the generated code is not directly executable in a single pass. For instance, models produce incomplete scripts or bugs, omit expected outputs, fail to parse data, or struggle with planning over geo-spatial data—ultimately requiring human intervention across multiple iterations. This limitation aligns with how we construct the oracle code in the benchmark. This issue is more severe in open-source models like Llama, which tend to produce fewer executable code. We, therefore, emphasize the need for stronger \textit{agentic} behaviors~\citep{plaat2025agentic, kapoor2024ai, Ng2024} in foundation models, where we define "agentic" as the ability to autonomously generate fully executable code for human end-users in a single interaction, particularly when the end-users are domain scientists rather than programmers.


\textbf{Fine-grained geo-spatial tasks reveals different strength-weakness tradeoffs}
Commercial and open-source models exhibit different strengths and weaknesses in fine-grained geo-spatial tasks, as shown in Figure~\ref{fig:results}. Specifically, open-source models generally struggle more than commercial ones in identifying regions with the most significant patterns. However, both types of models perform well when comparing trends between two locations or analyzing seasonal variations at a single location. In contrast, they show weaker performance when comparing seasonal variations across multiple locations or comparing data across different locations and times.

\textbf{Models perform better at identifying overall trends than fine-grained region detections} As mentioned in Figure~\ref{fig:overview}, target answers captures fine-grained aspects in answering these geo-spatial queries. Evaluation results in Figure~\ref{fig:results2} show that models perform best on the "trend" column, while accuracy drops for spatial, coordinate, or label references—highlighting a need for improvement in fine-grained regional understanding.

\textbf{Heatmaps with numerical annotations enhance performance, whereas map-overlaid heatmaps pose greater challenges for vision-language models}
Figure~\ref{fig:results3} compares model performance across three input image formats defined in Figure~\ref{fig:image_formats}. Adding numerical annotations to heatmaps improves model accuracy compared to using color gradients alone. In contrast, the most realistic format, where heatmaps are overlaid on geographic base maps, poses the greatest challenge for all models, as the added visual complexity hinders spatial pattern recognition. 

%% file: sections/related_work.tex
\section{Related Work}


\paragraph{Geo-Spatial Reasoning with LLMs}
Geo-spatial reasoning involves understanding and analyzing complex data based on its spatial and temporal relationships in the world~\citep{schottlander2025geospatial}. Most existing work focuses on Earth observation data from satellite or remote sensing imagery~\citep{lacoste2023geo, zhang2023geogpt, danish2024geobench, zhang2024good, zheng2023segment, wang2024earthvqa, muhtar2024lhrs, bazi2024rs, kuckreja2024geochat, tao2025advancements, liu2025gair}, performing scene understanding tasks such as object detection, semantic segmentation, object counting, captioning. Notable examples include GeoGPT~\citep{zhang2023geogpt}, GeoBench~\citep{danish2024geobench}, EarthVQA~\citep{wang2024earthvqa}, GEOBench-VLM~\citep{danish2024geobench}, and GeoChat~\citep{kuckreja2024geochat}.
However, gridded geo-spatial data—critical for capturing spatial and temporal patterns—remains largely overlooked in the AI-assisted geo-spatial research. Our work, \datasetname, specifically targets this gap by focusing on grid-based data in both tabular and image formats, and evaluating how foundation models can analyze the underlying patterns. Other efforts in geo-spatial research have focused on text-based data retrieval with tool usages, particularly through Geographic Information Systems (GIS)~\citep{nationalgeographic_gis_2025}, SQL, or GeoSPARQL~\citep{van2013open} queries~\citep{krechetova2025geobenchx, ning2025autonomous, mooney2023towards, li2023autonomous, jiang2024chatgpt, resch2025generative, zhang2023geogpt, jiang2024urbanllm} or Retrieval-Augmented Generation (RAG) systems~\citep{cromp2024climate, xie2024wildfiregpt, xie2025rag, vaghefi2023chatclimate, thulke2024climategpt, bulian2023assessing}. Representative works include GeoGPT~\citep{zhang2023geogpt}, GeoBenchX~\citep{krechetova2025geobenchx}, Autonomous GIS~\citep{li2023autonomous}, WildfireGPT~\citep{xie2024wildfiregpt, xie2025rag}, and ChatClimate~\cite{vaghefi2023chatclimate}. These approaches typically present geo-spatial information in textual formats and then rely on specific query syntax or semantic embeddings to interact with their databases. In contrast, our work sidesteps the data retrieval part and focuses on the geo-spatial data analysis itself. Furthermore, \citet{bhandari2023large} sets up basic geo-spatial knowledge, awareness, and reasoning tasks, such as querying the coordinates of a city, determining whether two locations are nearby, and predicting distances. \citet{mai2023opportunities} discusses geo-spatial semantics, geographical problem related to population health and urban planning, and remote sensing. GeoLLM~\citep{manvi2023geollm} focuses spatial information about locations.
Meanwhile, geo-spatial data also plays a central role in climate science, including WildfireGPT~\citep{xie2024wildfiregpt, xie2025rag}, ClimateGPT~\citep{thulke2024climategpt}, ChatClimate~\citep{vaghefi2023chatclimate}, and more~\citep{cromp2024climate, goecks2023disasterresponsegpt, chen2024optimizing, cao2024llm, hostetter2024large, martelo2024towards} that build on geo-spatial data in climate domains. Our work delves deeper into fine-grained trends and patterns across one or more data frames in multimodal, gridded representations.

\paragraph{Tabular Reasoning with LLMs}
Gridded geo-spatial data is often represented in tabular formats, posing unique challenges for language models in processing structured, numerically dense information. Current literature primarily focus on tables from databases with rich semantic annotations such as a descriptive name of each entity. Benchmarks like HybridQA, TabFact, ToTTo, WikiTQ, and others~\citep{chen2020hybridqa, chen2019tabfact, parikh2020totto,aly2021feverous, chen2021finqa, pasupat2015compositional} focus on simple fact extractions and \citet{he2024text2analysis, sui2024table} cover more advanced analysis that still rely heavily on semantic cues. In contrast, our work focuses on tables dominated by large volumes of numerical values, with spatial dependencies and no semantic annotations except for coordinates, presenting a different form of tabular reasoning~\citep{fang2024large, zhang2025survey}. To handle tabular data with language models, current work adopts strategies such as serializing tables into Markdown or other common formats~\citep{fang2024large, wang2024chain}, fine-tuning on tabular tasks~\citep{yang2023unitabe, zhang2023tablellama, li2023table, thomas2024retrieval}, leveraging tool use and code generation~\citep{fang2024large, cromp2024climate, cheng2022binding, zhang2023reactable}, or using image-based table representations~\citep{deng2024tables}. In our work, we extend this line of research by visualizing tables with geo-spatial semantics heatmaps or overlays on actual maps and by exploring code-based analysis in geo-spatial contexts that introduce unique challenges.

%% file: sections/conclusion.tex
\section{Conclusion}
We introduced \datasetnamelogo, a comprehensive benchmark designed to evaluate the capability of foundation models to understand 
multimodal gridded geo-spatial data. 
\datasetname~features structured, dense numerical data using real-world gridded datasets and expert-curated templates to evaluate scientifically relevant geo-spatial tasks.
This integrated design enables robust and scalable assessment of foundation models across vision, language, and code modalities.
Our evaluation reveals that while vision-language models excel at interpreting spatial patterns from heatmaps, they still struggle with fine-grained regional understanding and label-based reasoning. Meanwhile, language and code models show limited success in generating executable analysis scripts without human intervention, highlighting the need for stronger agentic behavior. These findings point to several critical areas where model capabilities must improve to meet the practical needs of geo-spatial scientific analysis. Overall, this work can inform the development of more capable models to process and understand the dense numerical data, spatiotemporal dependencies, and multimodal representations of geo-spatial data, supporting the advancement of foundation models for informed decision-making and resilience building across a wide range of real-world challenges.


\textbf{Limitations and Future Work}
\label{sec: limitation}
We acknowledge that this dataset is limited to the United States due to data availability. Additionally, our benchmark focuses on geo-spatial data in gridded formats, intentionally excluding other common data types such as Earth observation and remote sensing imagery, which have already been extensively studied in prior work. However, the underlying framework are designed to be generalizable and can be readily applied to similar gridded geo-spatial datasets from other regions. Building on this foundation, future work will focus on expanding \datasetname~beyond the United States and incorporating richer data modalities such as satellite imagery, elevation maps, and land use data to enable broader and more diverse analytical capabilities.

%% file: sections/appendix.tex
\newpage
\appendix

\section{Inference Prompts}\label{sec:prompt}
To evaluate models across different modalities, we design prompts for three settings: language-only, language and code, and language and vision. Each prompt is designed to be simple yet encourage model response with desired style and consistent answer formatting.

\begin{itemize}
    \item Language-only: models receive data in tabular format with instructions \textit{"Think step by step before making a decision. Then, explicitly state your final choice after the special phrase "\#\#\#\#Final Answer" followed by (a), (b), (c), or (d). Please don't use programming code."}.
    
    \item Language and programming code: models receive data in tabular format with instructions \textit{"Please write Python code to answer the question and show the complete script. You must include a print statement at the end of the code that outputs the final answer using the special phrase "\#\#\#\#Final Answer' followed by (a), (b), (c), or (d)."}
    
    \item Language and vision: models receive climate data in one of the three image formats with instructions \textit{"Analyze this image and answer the question. Think step by step before making a decision. Then, explicitly state your final choice after the special phrase "\#\#\#\#Final Answer" followed by (a), (b), (c), or (d)."}.
\end{itemize}

\section{Examples of Data Visualizations for All Query Templates}\label{sec:more_vis}

\begin{figure}[!ht]
  \centering
  \includegraphics[width=0.33\linewidth,height=0.8\textheight,keepaspectratio]{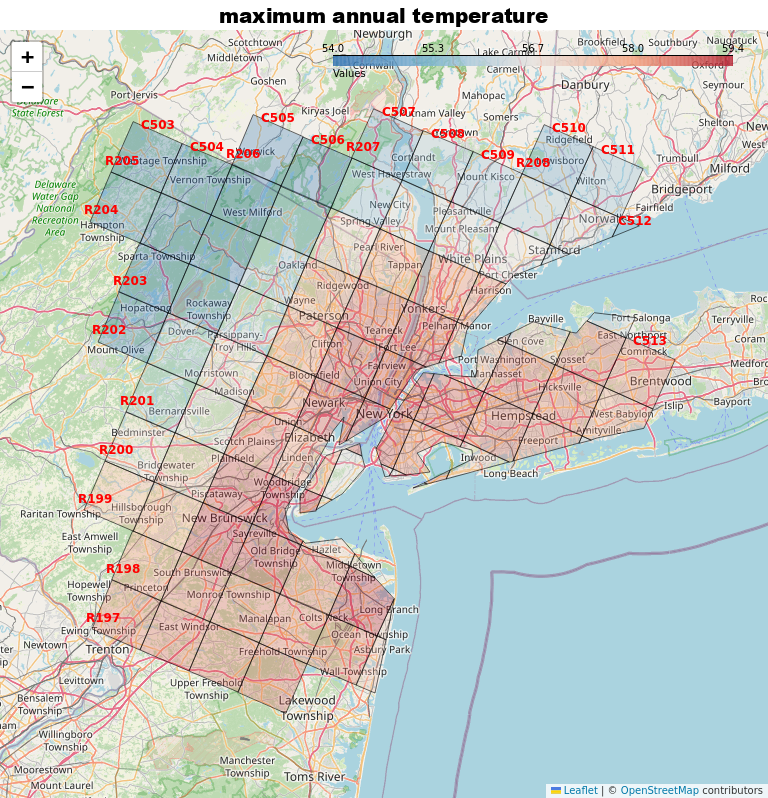}\hfill
  \includegraphics[width=0.33\linewidth,height=0.8\textheight,keepaspectratio]{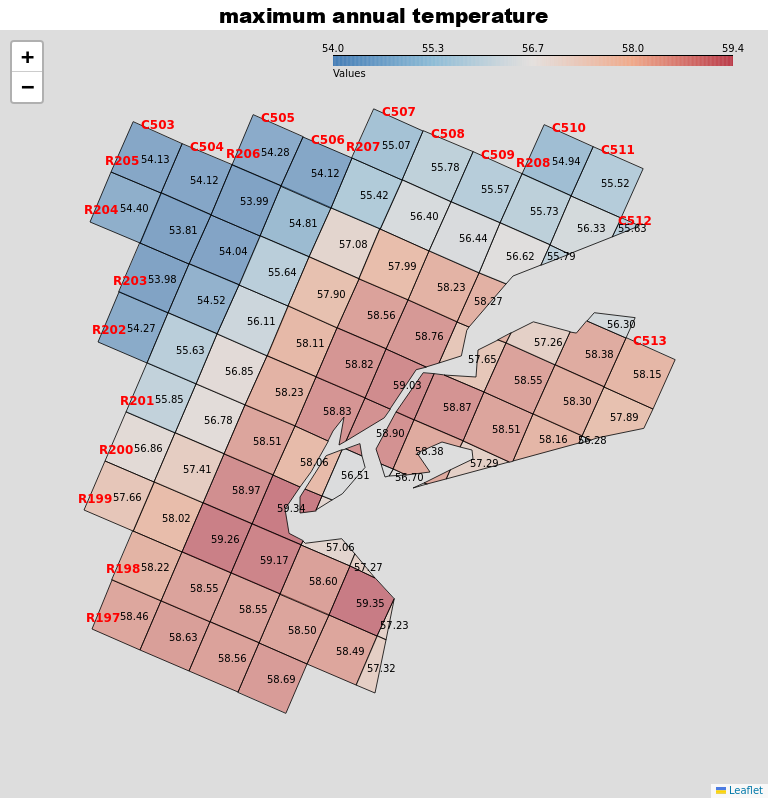}\hfill
  \includegraphics[width=0.33\linewidth,height=0.8\textheight,keepaspectratio]{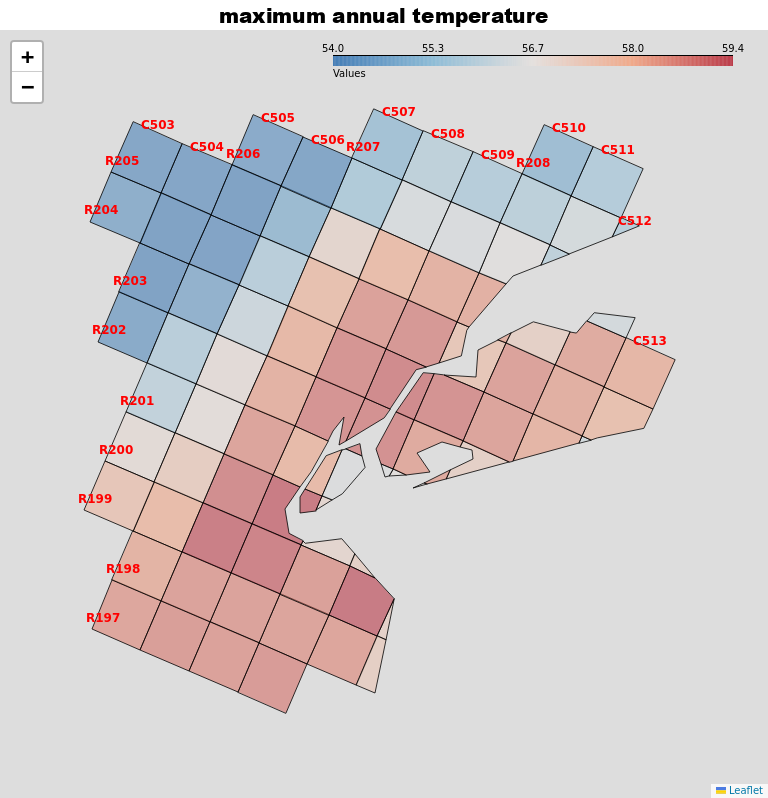}
  \caption{\textbf{Template 1:} Which region in \{{location1}\} experienced the largest increase in \{{climate\_variable1}\} during \{{time\_frame1}\}? This example takes location1 = New York city, NY, climate\_variable1 = maximum annual temperate, and time\_frame1 = historical period.}
  \label{fig:template1}
\end{figure}

\begin{figure}[!ht]
  \centering
  \includegraphics[width=0.8\linewidth,height=0.5\textheight,keepaspectratio]{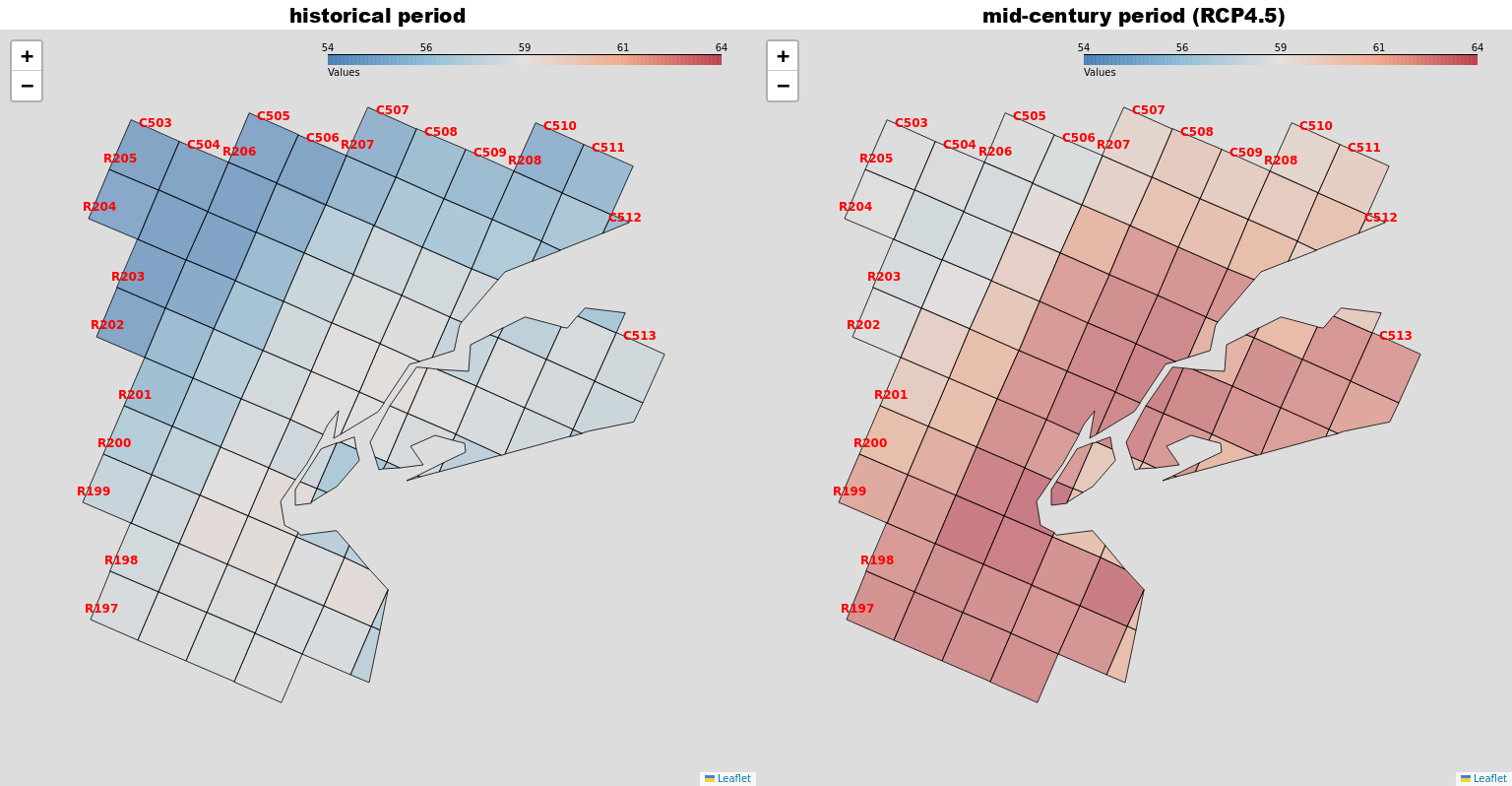}\\[1ex]
  \includegraphics[width=0.8\linewidth,height=0.5\textheight,keepaspectratio]{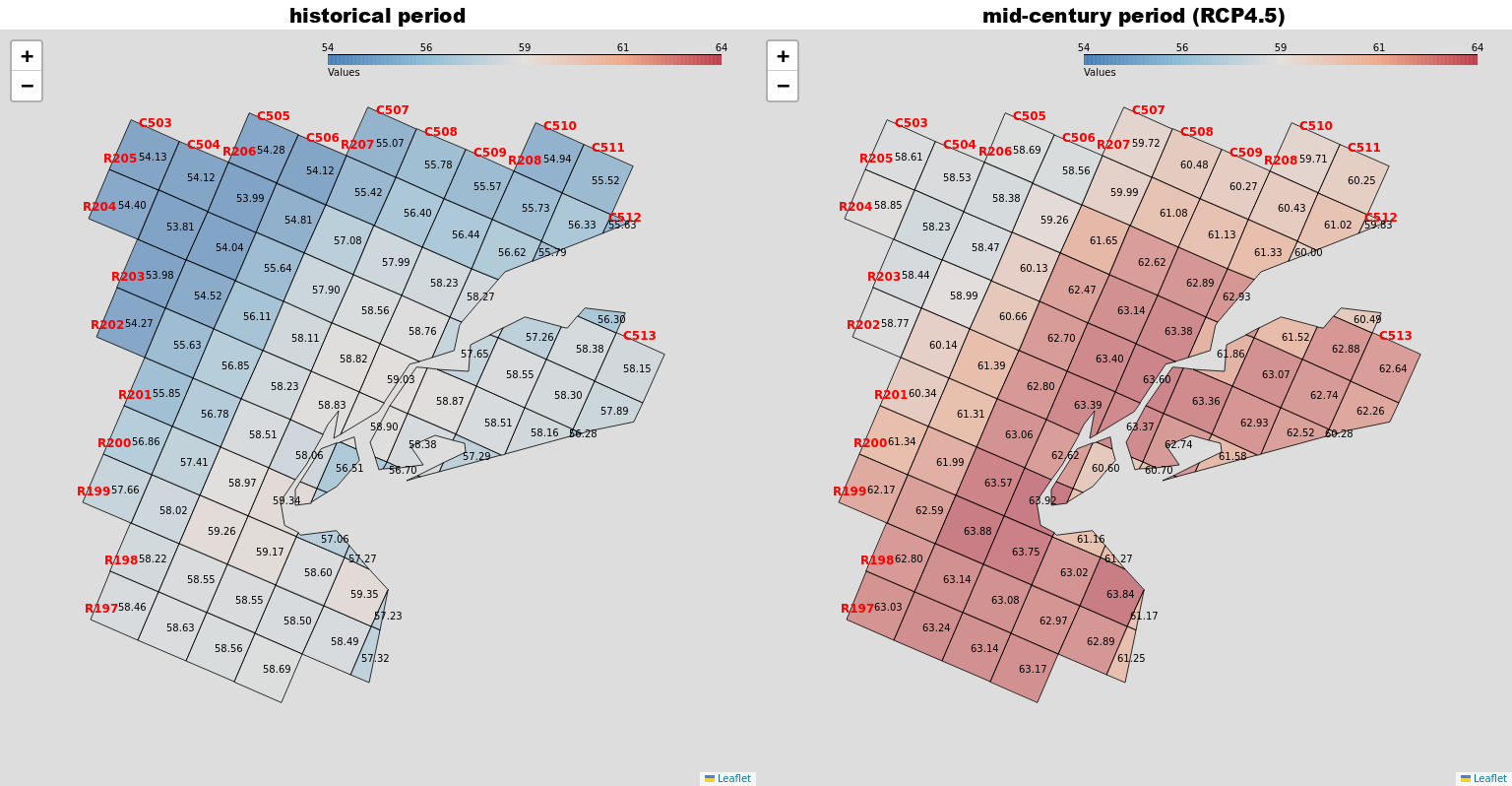}\\[1ex]
  \includegraphics[width=0.8\linewidth,height=0.8\textheight,keepaspectratio]{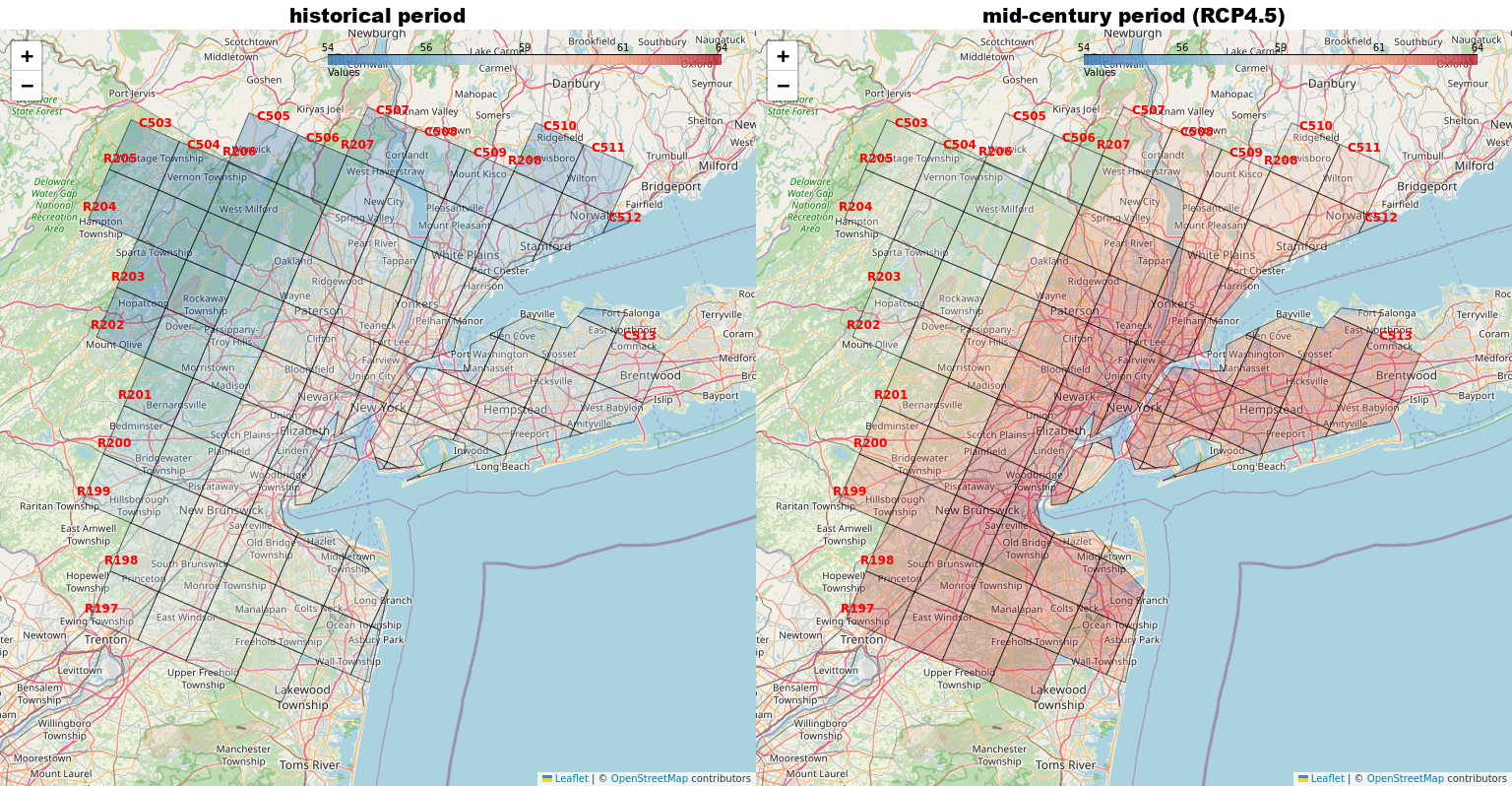}
  \caption{\textbf{Template 2:} How has \{{climate\_variable1}\} changed between \{{time\_frame1}\} and \{{time\_frame2}\} in the \{{location1}\}? This example takes location1 = New York city, NY, climate\_variable1 = maximum annual temperate, time\_frame1 = historical period, and time\_frame2 = mid-century period (RCP-4.5).}
  \label{fig:template2}
\end{figure}

\begin{figure}[!ht]
  \centering
  \includegraphics[width=0.8\linewidth,height=0.8\textheight,keepaspectratio]{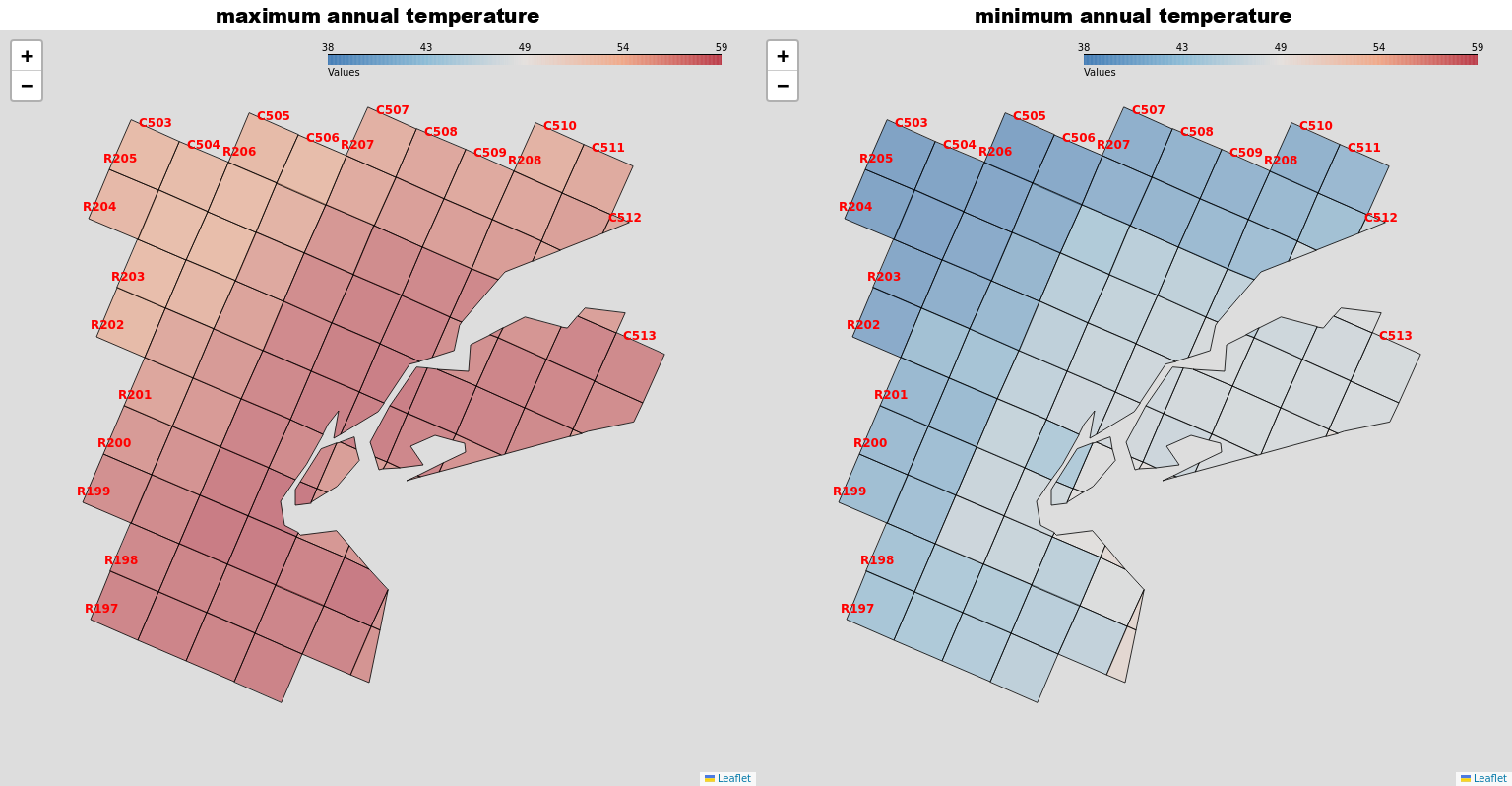}\\[1ex]
  \includegraphics[width=0.8\linewidth,height=0.8\textheight,keepaspectratio]{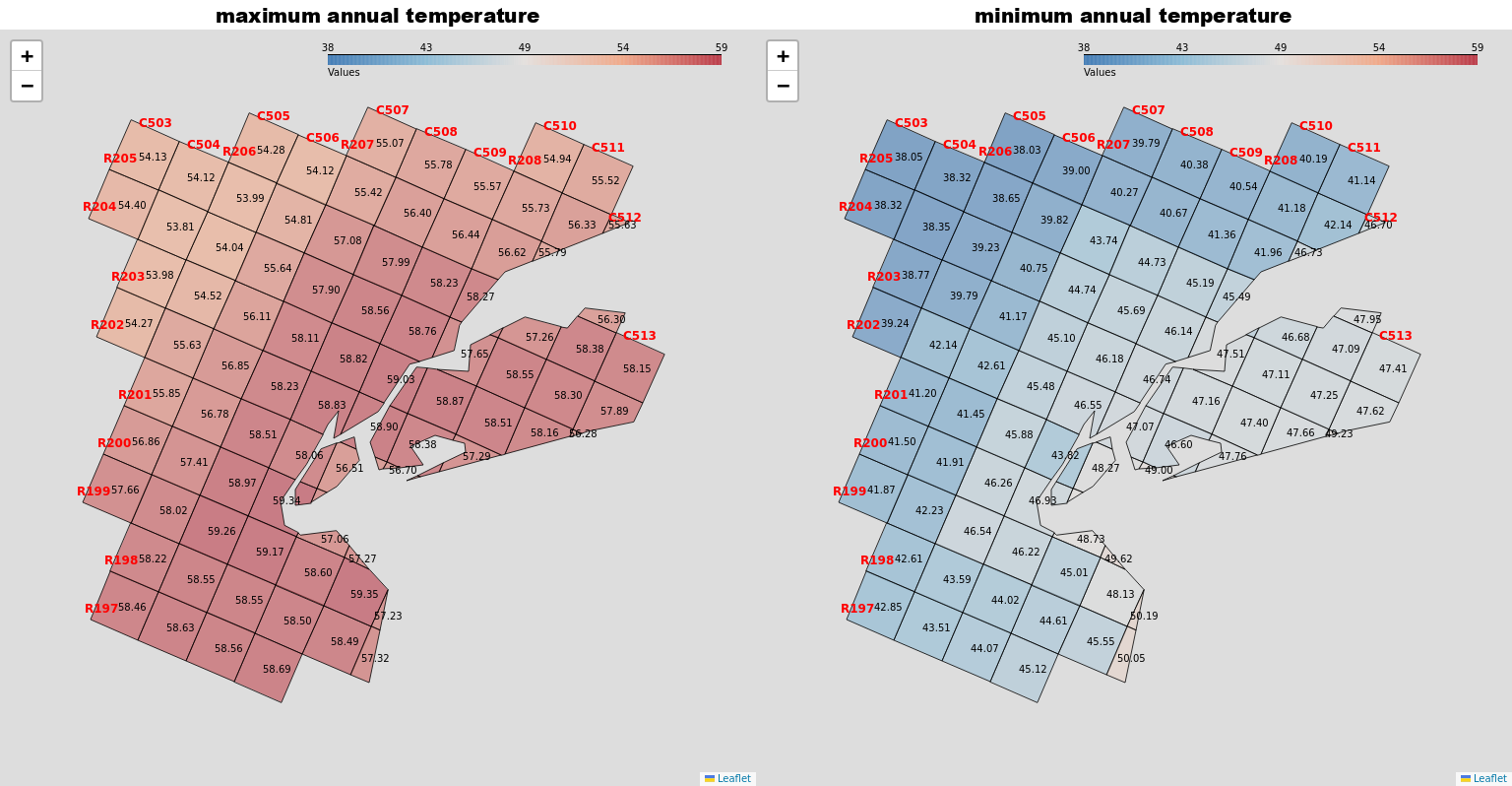}\\[1ex]
  \includegraphics[width=0.8\linewidth,height=0.8\textheight,keepaspectratio]{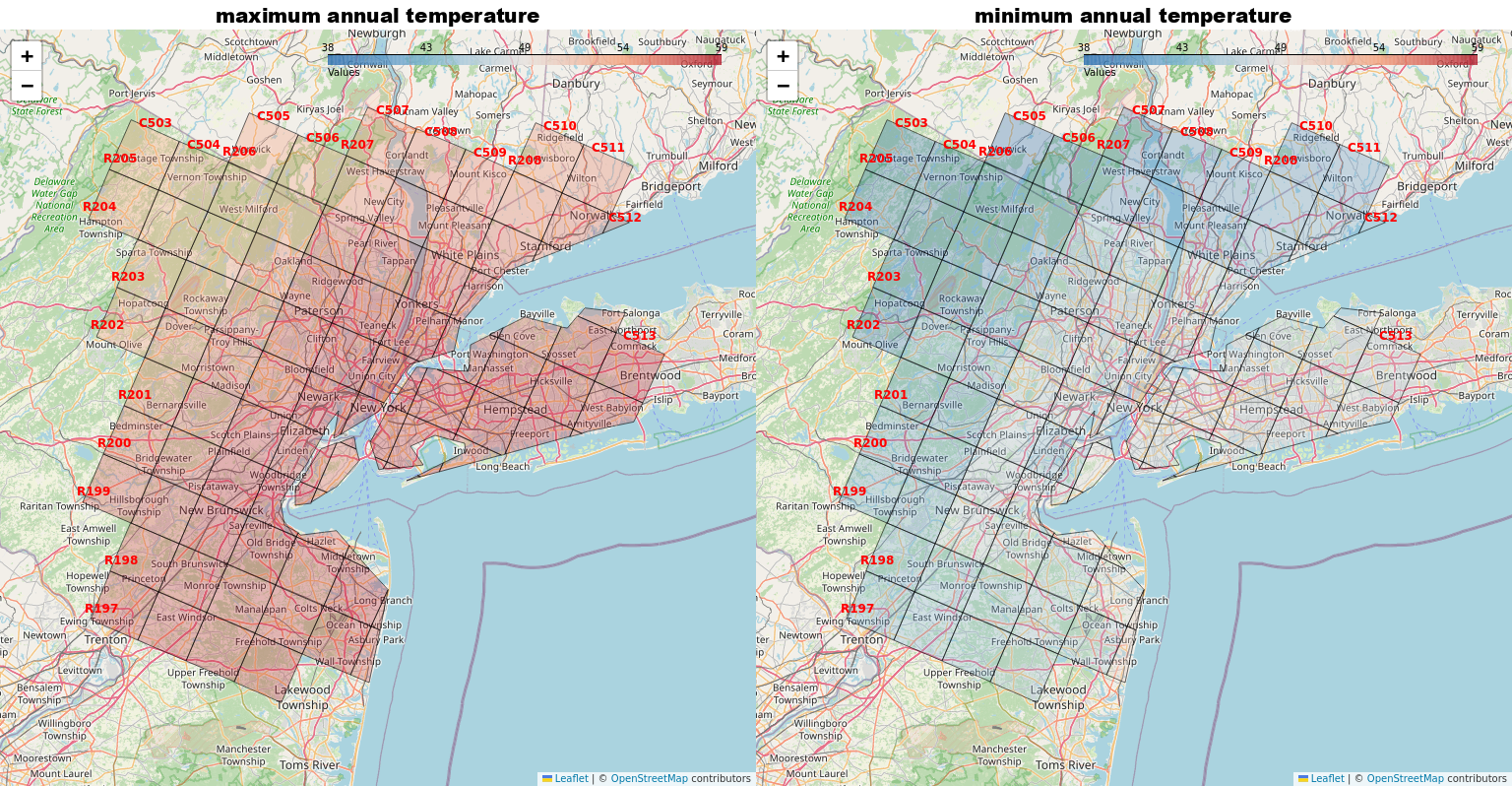}
  \caption{\textbf{Template 3:} What is the correlation between \{{climate\_variable1}\} and \{{climate\_variable2}\} in the \{{location1}\} during \{{time\_frame1}\}? This example takes location1 = New York city, NY, climate\_variable1 = maximum annual temperate, climate\_variable2 = minimum annual temperate, and time\_frame1 = historical period.}
  \label{fig:template3}
\end{figure}

\begin{figure}[!ht]
  \centering
  \includegraphics[width=0.8\linewidth,height=0.8\textheight,keepaspectratio]{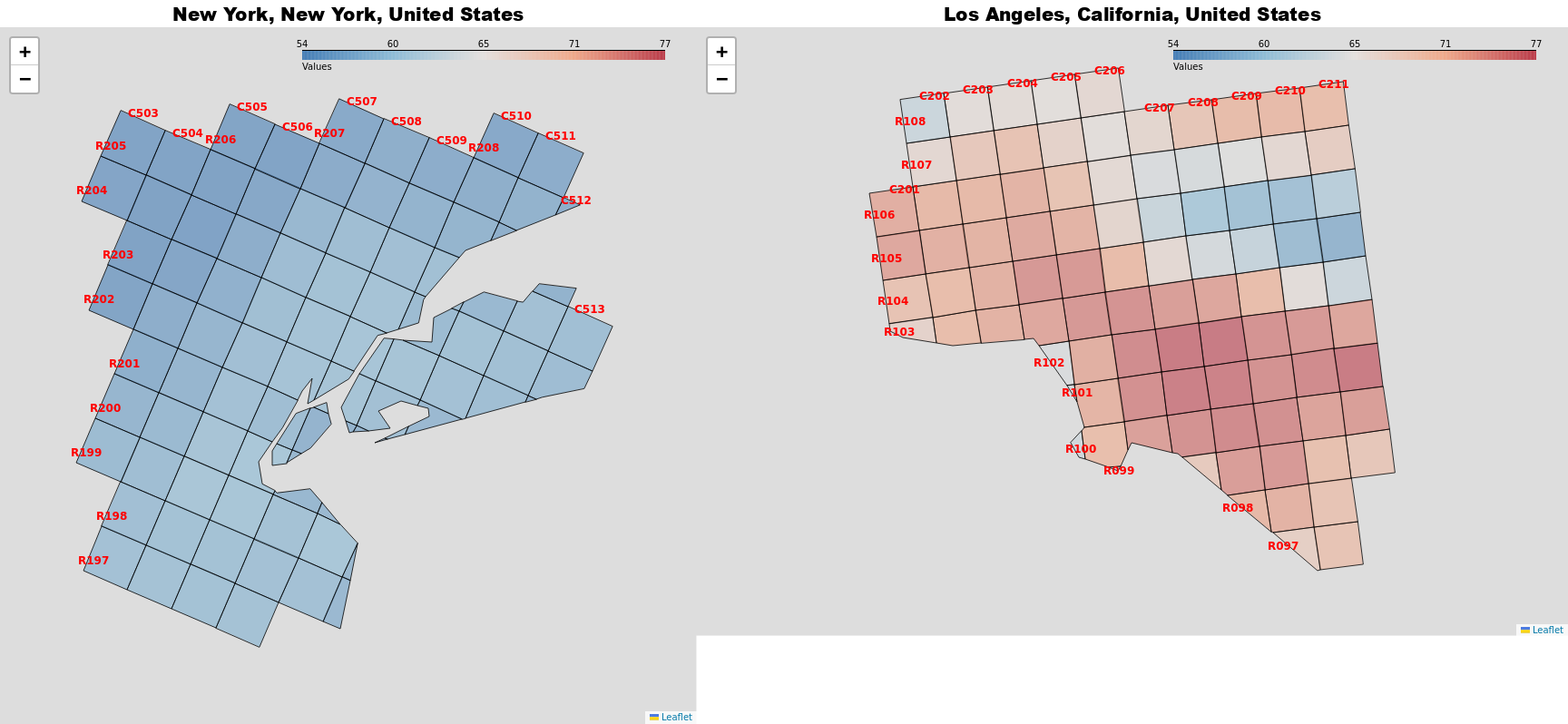}\\[1ex]
  \includegraphics[width=0.8\linewidth,height=0.8\textheight,keepaspectratio]{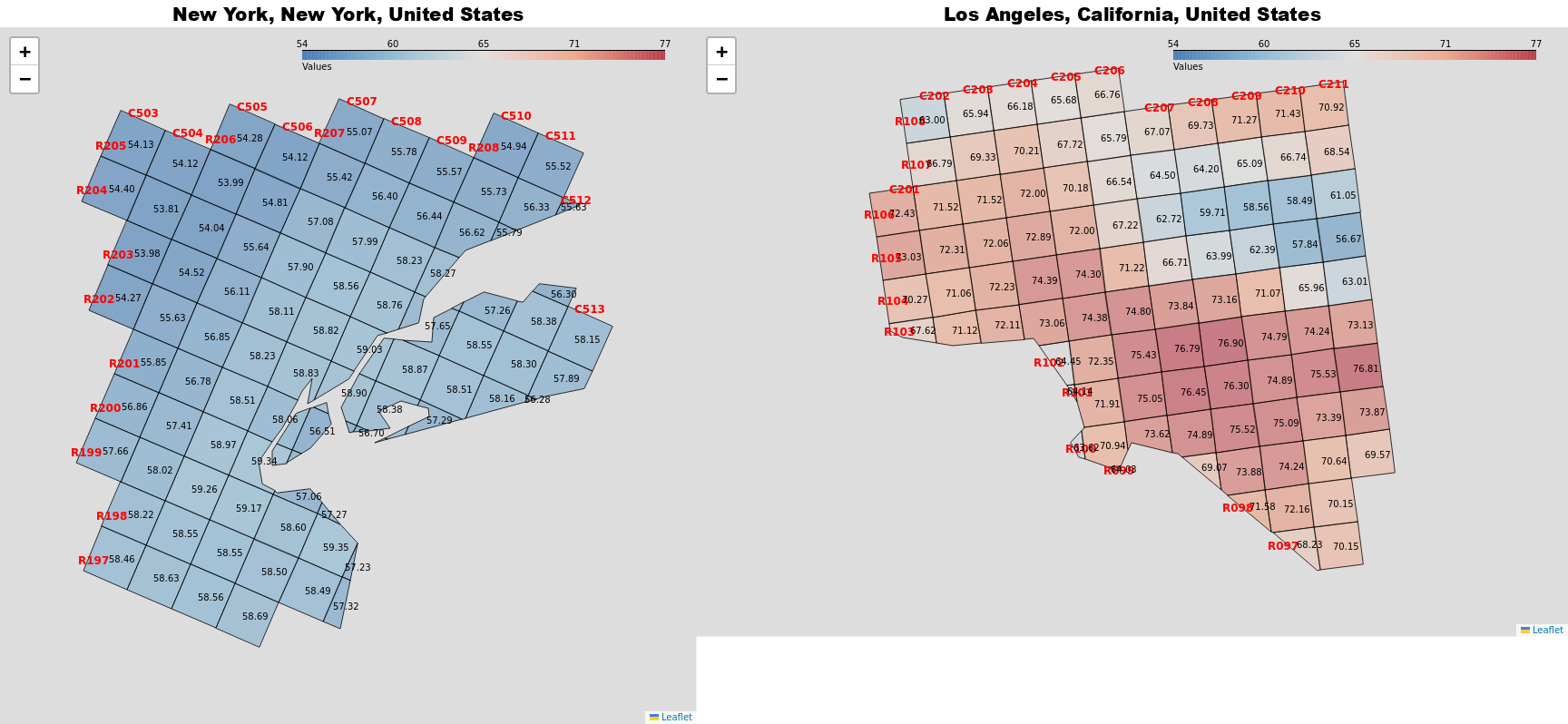}\\[1ex]
  \includegraphics[width=0.8\linewidth,height=0.8\textheight,keepaspectratio]{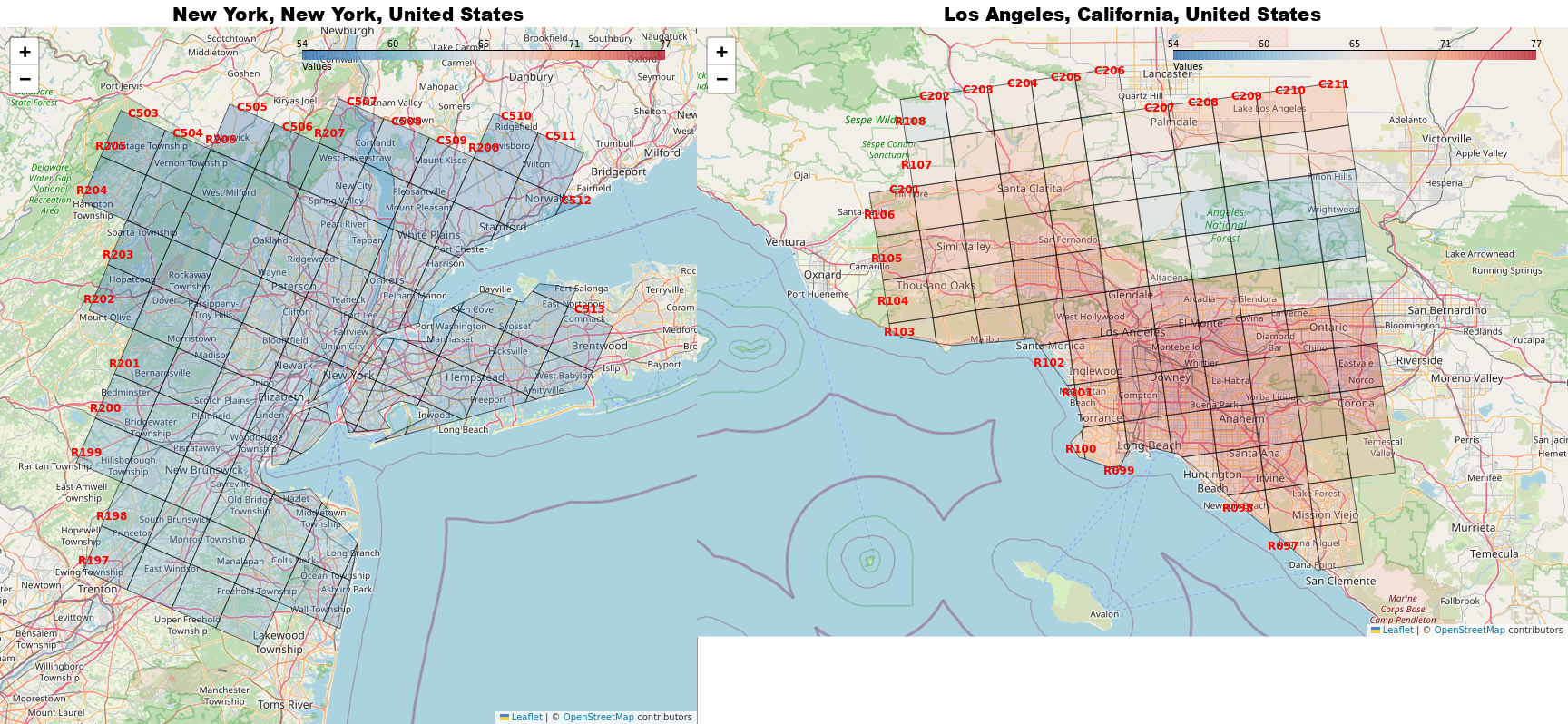}
  \caption{\textbf{Template 4:} How does \{{climate\_variable1}\} compare between \{{location1}\} and \{{location2}\} during \{{time\_frame1}\}? This example takes location1 = New York city, NY, location2 = Los Angeles, CA, climate\_variable1 = maximum annual temperate, and time\_frame1 = historical period.}
  \label{fig:template4}
\end{figure}

\begin{figure}[!ht]
  \centering
  \includegraphics[width=\linewidth,height=0.8\textheight,keepaspectratio]{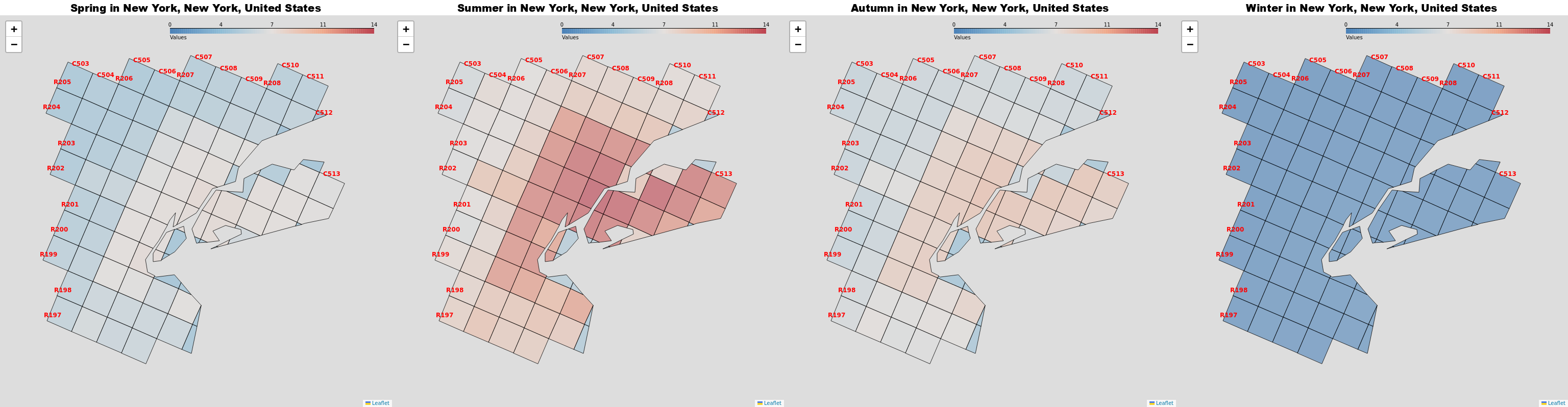}\\[1ex]
  \includegraphics[width=\linewidth,height=0.8\textheight,keepaspectratio]{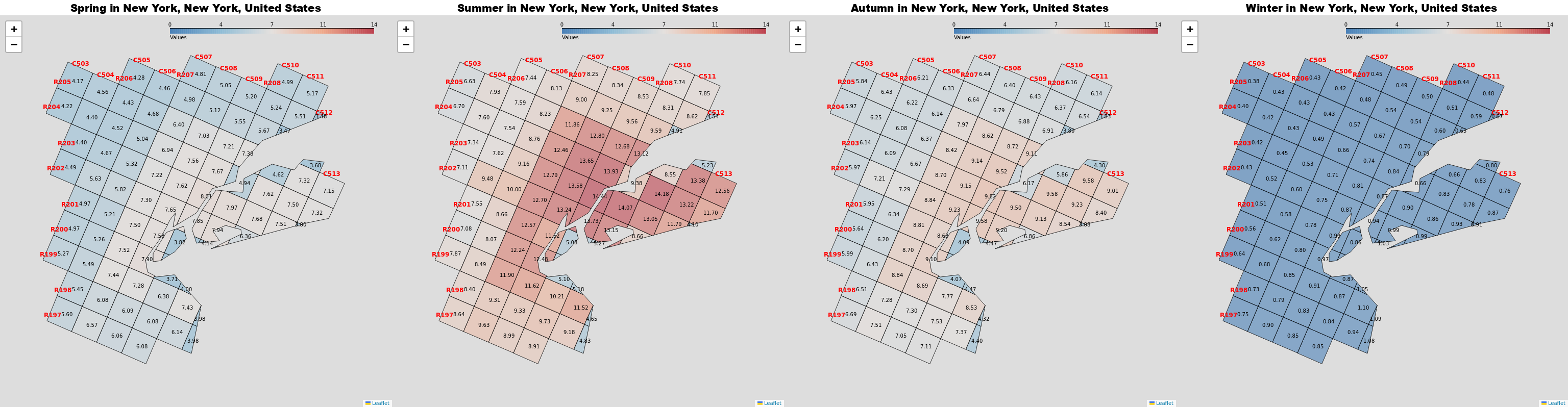}\\[1ex]
  \includegraphics[width=\linewidth,height=0.8\textheight,keepaspectratio]{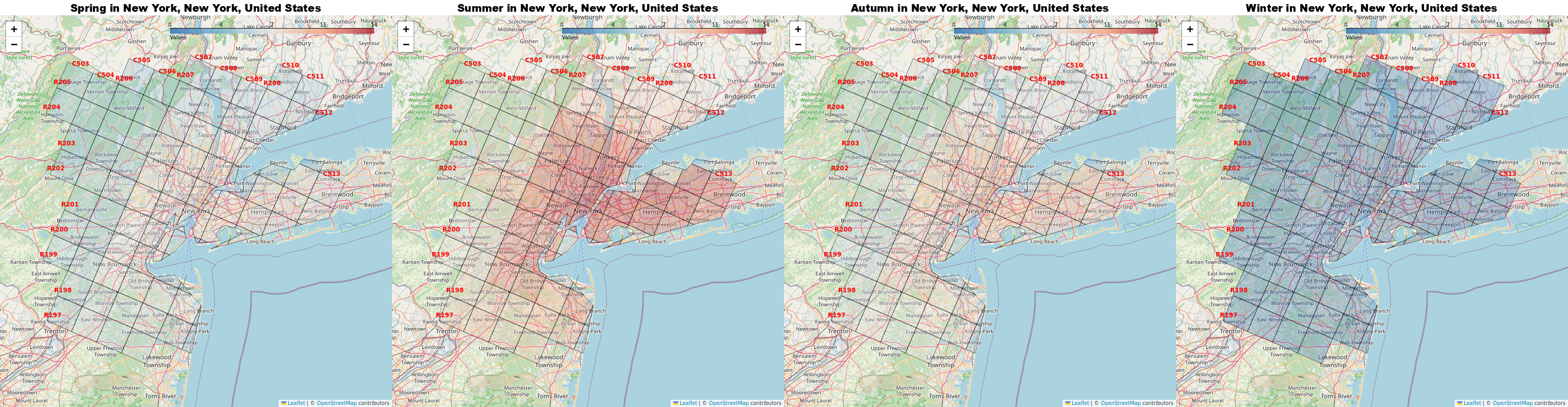}
  \caption{\textbf{Template 5}: What is the \emph{seasonal} variation of \{{climate\_variable1}\} in \{{location1}\} during \{{time\_frame1}\}? Same data is used in \textbf{Template 6}: Which \emph{season} in \{{time\_frame1}\} saw the highest levels of \{{climate\_variable1}\} in \{{location1}\}? This example takes location1 = New York city, NY, climate\_variable1 = maximum annual temperate, and time\_frame1 = historical period.}
  \label{fig:template5}
\end{figure}

\begin{figure}[!ht]
  \centering
  \includegraphics[width=0.6\linewidth,height=0.8\textheight,keepaspectratio]{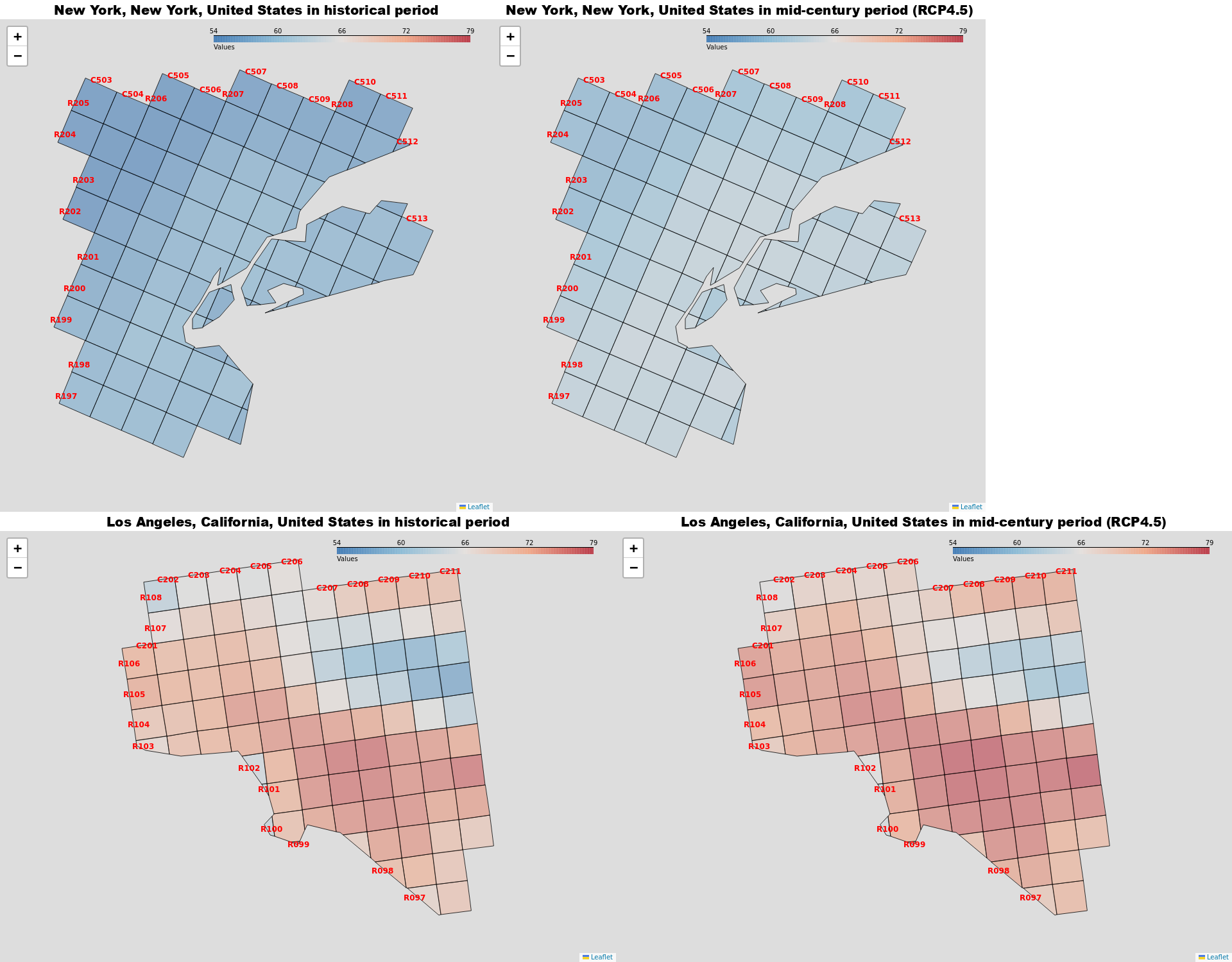}\\[1ex]
  \includegraphics[width=0.6\linewidth,height=0.8\textheight,keepaspectratio]{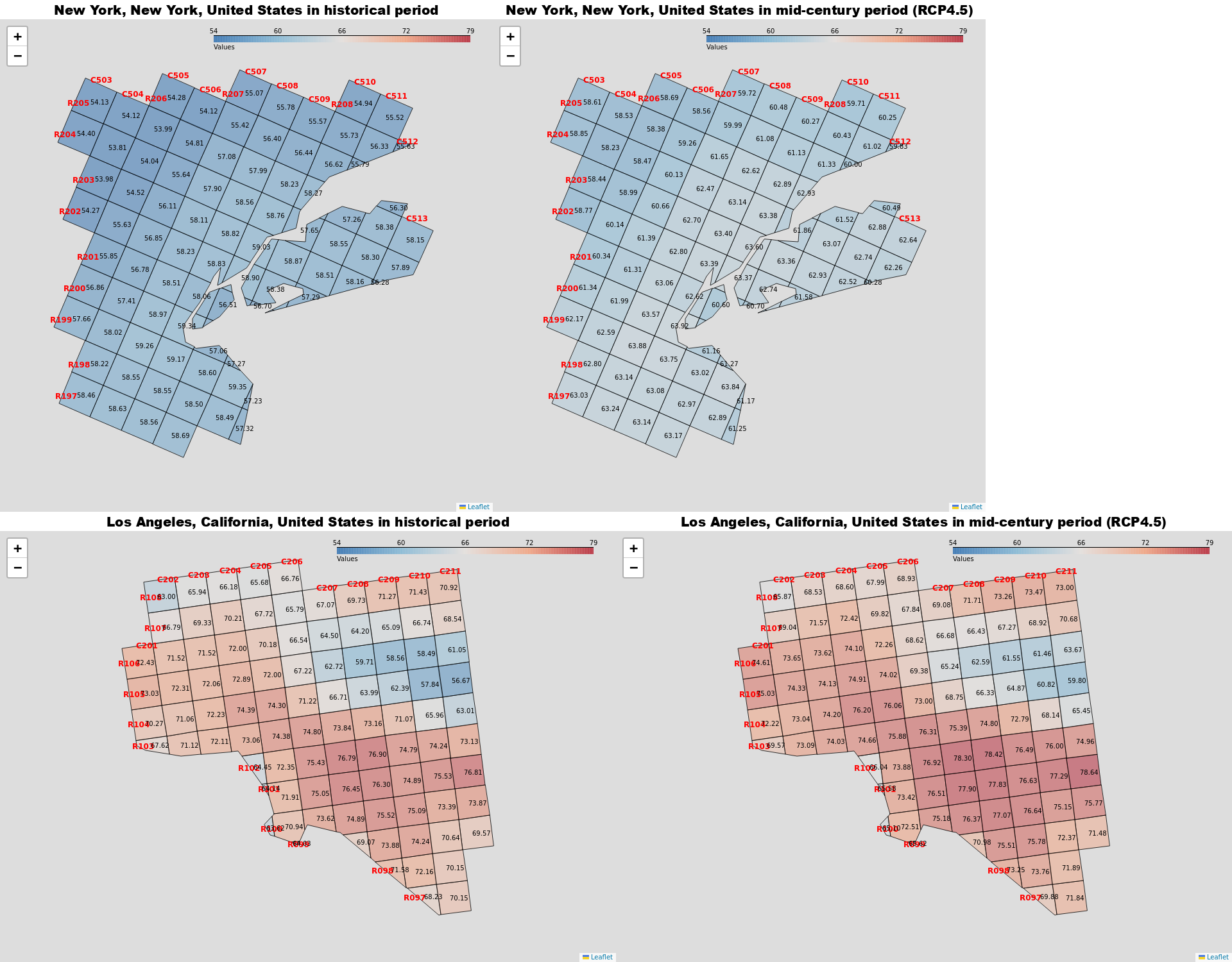}\\[1ex]
  \includegraphics[width=0.6\linewidth,height=0.8\textheight,keepaspectratio]{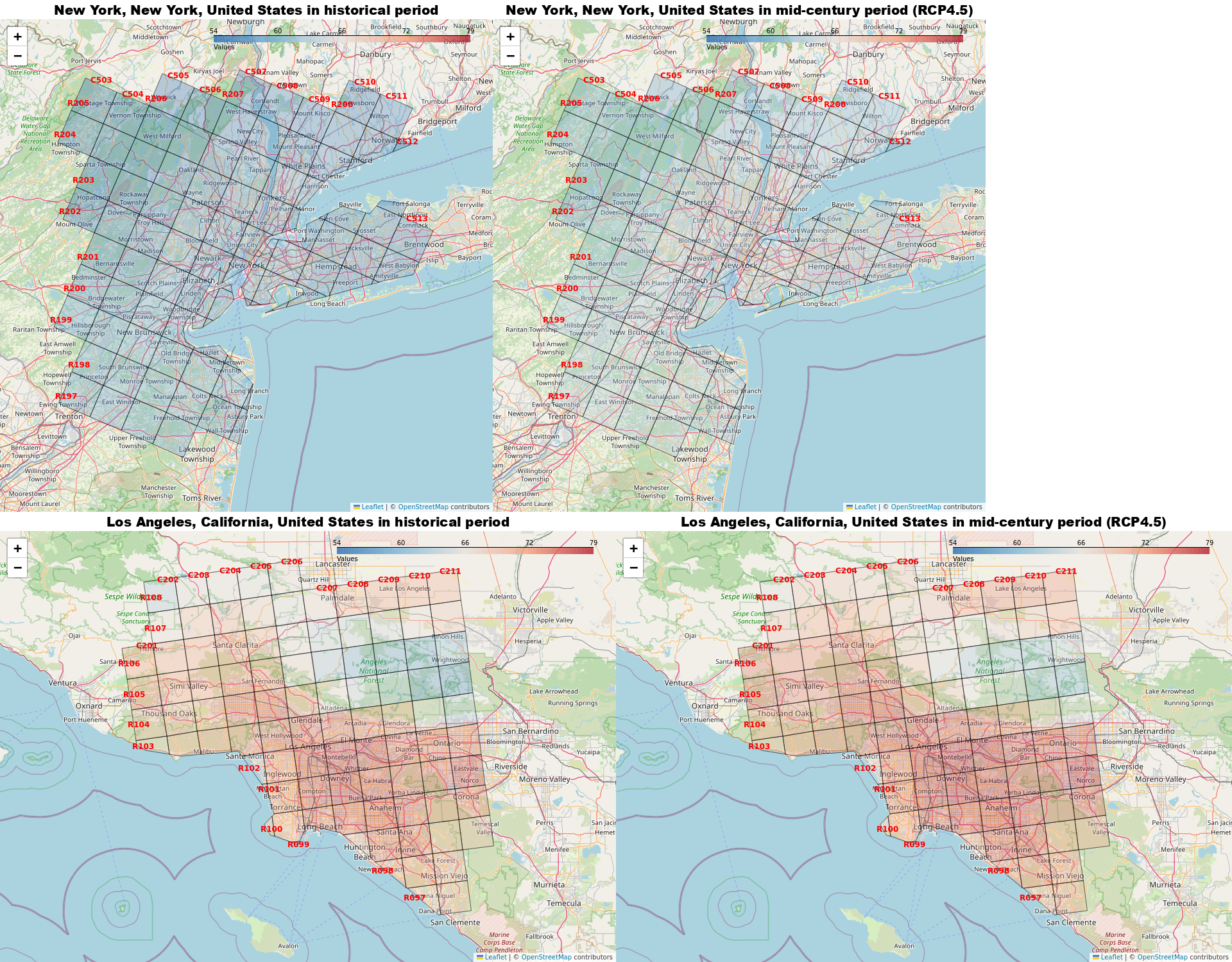}
  \caption{\textbf{Template 7.} Which of \{{location1}\} or \{{location2}\} experienced a greater change in \{{climate\_variable1}\} throughout \{{time\_frame1}\} and \{{time\_frame2}\}? This example takes location1 = New York city, NY, location2 = Los Angeles, CA, climate\_variable1 = maximum annual temperate, time\_frame1 = historical period, and time\_frame1 = mid-century period (RCP4.5).}
  \label{fig:template7}
\end{figure}

\begin{figure}[!ht]
  \centering
  \includegraphics[width=\linewidth,height=0.8\textheight,keepaspectratio]{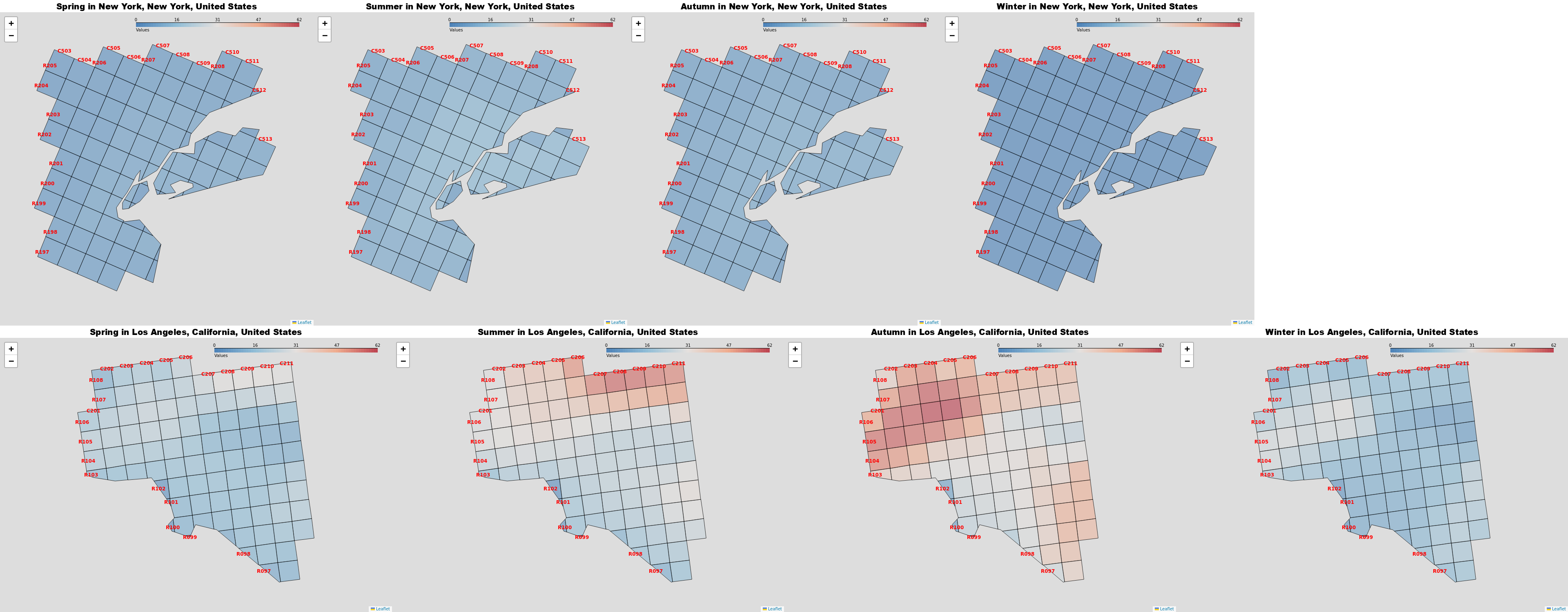}\\[1ex]
  \includegraphics[width=\linewidth,height=0.8\textheight,keepaspectratio]{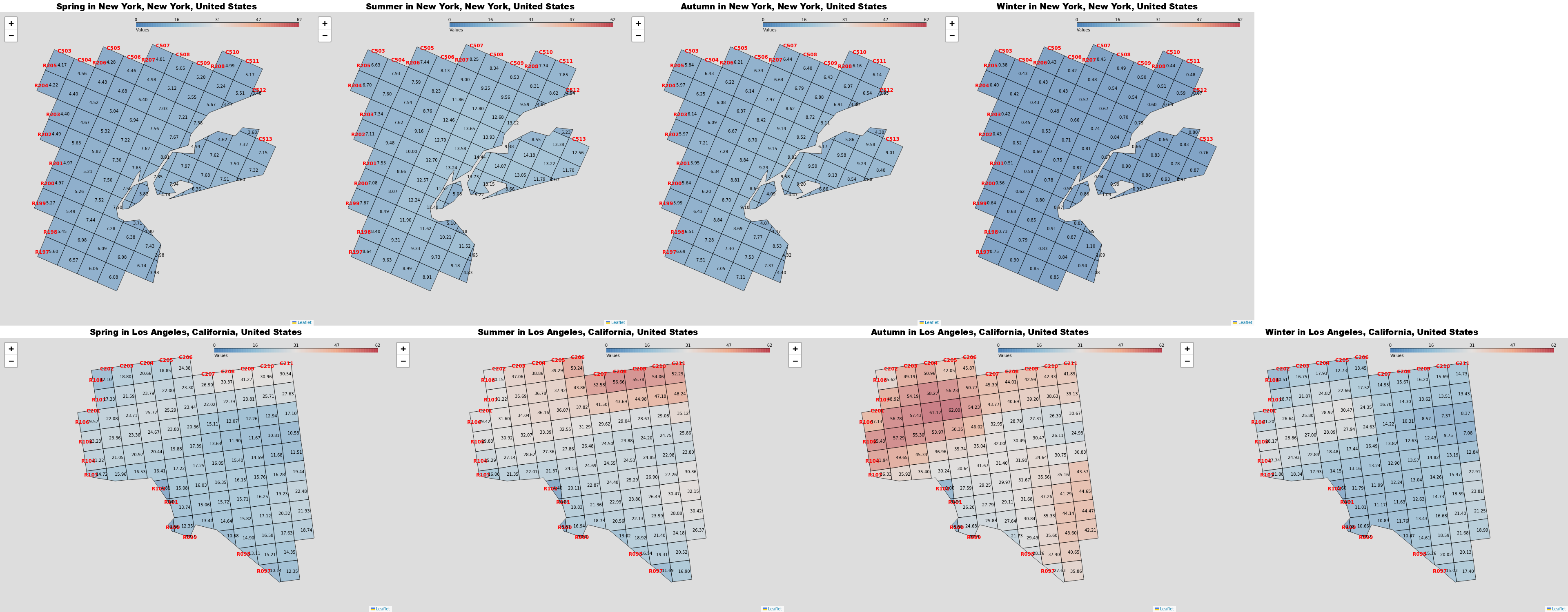}\\[1ex]
  \includegraphics[width=\linewidth,height=0.8\textheight,keepaspectratio]{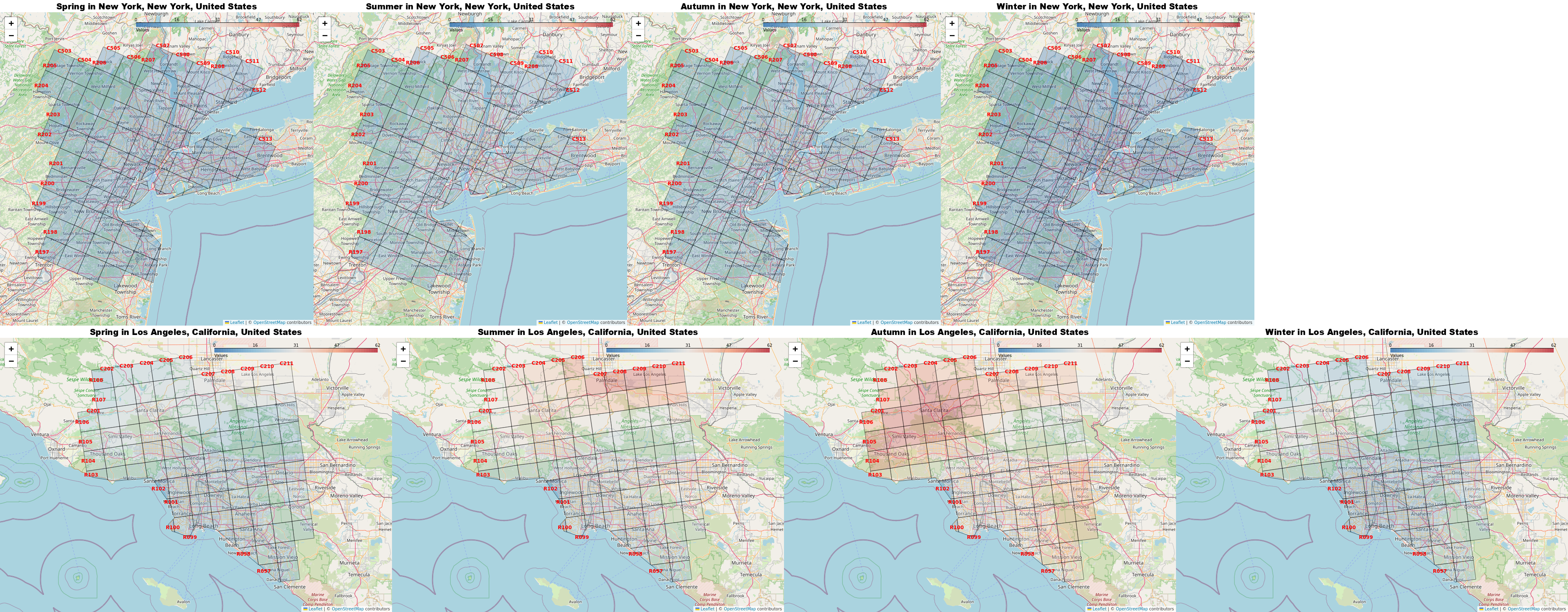}
  \caption{\textbf{Template 8.} How does the \emph{seasonal} variation of \{{climate\_variable1}\} in \{{location1}\} compare to that in \{{location2}\} for \{{time\_frame1}\}? This example takes location1 = New York city, NY, climate\_variable1 = maximum annual temperate, and time\_frame1 = historical period.}
  \label{fig:template8}
\end{figure}
\FloatBarrier